\pdfoutput=1

\documentclass[acmsmall, nonacm]{acmart}

\usepackage{pifont}
\newcommand{\cmark}{\ding{51}}

\usepackage{rotating}
\usepackage{multirow}
\usepackage{subcaption}
\AtBeginDocument{%
  }

\setcopyright{none}
\copyrightyear{2025}
\acmYear{nonacm}
\acmDOI{XXXXXXX.XXXXXXX}
\acmVolume{X}
\acmNumber{X}
\acmArticle{X}
\acmMonth{X}

\begin{document}

\title{System Log Parsing with Large Language Models: A Review}


\author{Viktor Beck}
\affiliation{%
  \institution{AIT Austrian Institute of Technology}
  \city{Vienna}
  \country{Austria}}
\email{viktor.beck@ait.ac.at}
\author{Max Landauer}
\affiliation{%
  \institution{AIT Austrian Institute of Technology}
  \city{Vienna}
  \country{Austria}}
\email{max.landauer@ait.ac.at}
\author{Markus Wurzenberger}
\affiliation{%
  \institution{AIT Austrian Institute of Technology}
  \city{Vienna}
  \country{Austria}}
\email{markus.wurzenberger@ait.ac.at}
\author{Florian Skopik}
\affiliation{%
  \institution{AIT Austrian Institute of Technology}
  \city{Vienna}
  \country{Austria}}
\email{florian.skopik@ait.ac.at}

\author{Andreas Rauber}
\affiliation{%
  \institution{TU Wien}
  \city{Vienna}
  \country{Austria}}
\email{andreas.rauber@tuwien.ac.at}

\renewcommand{\shortauthors}{Beck et al.}

\begin{abstract}
Log data provides crucial insights for tasks like monitoring, root cause analysis, and anomaly detection. Due to the vast volume of logs, automated log parsing is essential to transform semi-structured log messages into structured representations. Recent advances in large language models (LLMs) have introduced the new research field of LLM-based log parsing. Despite promising results, there is no structured overview of the approaches in this relatively new research field with the earliest advances published in late 2023. This work systematically reviews 29 LLM-based log parsing methods. We benchmark seven of them on public datasets and critically assess their comparability and the reproducibility of their reported results. Our findings summarize the advances of this new research field, with insights on how to report results, which data sets, metrics and which terminology to use, and which inconsistencies to avoid, with code and results made publicly available for transparency.
\end{abstract}



\maketitle

\section{Introduction}\label{sec:intro}

Log data consists of recorded information generated by logging facilities in software applications, systems, devices, or networks during operation. It provides a chronological record of events and contains valuable insights for various tasks, such as monitoring \cite{chen_experience_2019, wang_would_2021}, root cause analysis \cite{he_survey_2021, notaro_logrule_2023}, and anomaly detection \cite{landauer_deep_2023, he_experience_2016}. Given that modern computer systems can generate millions of log lines per hour, manual processing is infeasible, necessitating fast and scalable automated approaches. For instance, in 2013 the Alibaba Cloud system was reported to produce approximately 100 to 200 million log lines per hour \cite{mi_toward_2013}. However, extracting meaningful information from raw log data requires transforming it into a structured format. Since logs are typically semi-structured, following general patterns but containing variable log messages, direct analysis is challenging. To enable log analytics tasks like anomaly detection, logs must first be parsed into a structured representation, making log parsing a critical step in log analysis \cite{wurzenberger_aecid-pg_2019}.

Log parsing refers to the extraction of structured data from semi-structured logs. Parsing the logs into the structured format can simply be achieved by regular expression matching if the correct log templates are given. Most methods therefore formulate log parsing as a log template extraction problem \cite{he_drain_2017, jiang_lilac_2024, vaarandi_using_2024, yu_brain_2023, sedki_effective_2022, dai_logram_2022}. Ground-truth templates are defined somewhere in logging statements in the source code and are therefore often not available. A wide range of log parsing techniques \cite{shima_length_2016, jiang_abstracting_2008, du_spell_2016, sedki_effective_2022, le_log_2023-1} have therefore been proposed to overcome this issue.

Many state-of-the-art log parsing methods require manual configuration or other manual actions at some point in the parsing process. Approaches may involve labeling of some of the logs from the training set \cite{jiang_lilac_2024, xu_divlog_2024, le_log_2023-1}, definition of log formats \cite{li_did_2023} or regular expressions for preprocessing \cite{yu_brain_2023, he_drain_2017, sedki_effective_2022}, or other parameters \cite{yu_brain_2023} that fit the parsing algorithm to the data. From a user's perspective the configuration of parsing algorithms might seem overwhelming since it requires expertise or an analysis of the data at hand. The reason log parsing methods require manual actions is that humans have both semantic and syntactic understanding \cite{jiang_lilac_2024}, but more importantly, a natural generalization capability and potentially expert knowledge about logs and log parsing. Many log parsers have already achieved semantic \cite{le_log_2023-1, liu_uniparser_2022} and syntactic abilities \cite{he_drain_2017, dai_logram_2022, yu_brain_2023}, but many do not achieve satisfactory performance on data other than LogHub \cite{zhu_loghub_2023} as reported by \cite{fu_investigating_2022, zhang_eclipse_2024, li_revisiting_2024} or require substantial amounts of labeled logs for training their models \cite{xiao_demonstration-free_2024}. While syntax-based parsers focus on characteristics like occurrence frequencies, or word or log length, to determine the static parts of logs, semantic-based parsers leverage the semantic differences between static and variable parts of log messages \cite{jiang_lilac_2024, xiao_demonstration-free_2024}. Knowledge about logs and how to parse them can be learned by language models, such as BERT \cite{devlin_bert_2019} or its descendants, by fine-tuning them with logs and their templates \cite{le_log_2023-1, tao_logstamp_2022, liu_uniparser_2022}. However, these methods often lack generalization and have to be trained or fine-tuned \cite{gao_making_2021} with labeled logs, preferably with logs of the exact same log type as the target logs \cite{ma_llmparser_2024}. Generalization arises from pretraining on a large corpus of diverse datasets, allowing models to perform well also on novel tasks and possibly enhanced through fine-tuning \cite{gao_making_2021} or in-context learning (ICL) \cite{brown_language_2020}. This is where large language models (LLMs) come into play.

With the emergence of LLMs the new research field of LLM-based log parsing arised. In late 2023, Le and Zhang \cite{le_log_2023} reported notable performance of ChatGPT\footnote{\url{https://openai.com/index/chatgpt/}; accessed 10-March-2025} in a naive setting, where the LLM was presented individual logs and simply asked for their templates. In the meantime, other approaches adopted LLMs for this purpose as well, building sophisticated frameworks that utilize LLMs in various ways to enhance log parsing by adopting learning paradigms for LLMs such as ICL \cite{brown_language_2020} or fine-tuning \cite{gao_making_2021}. Given the fact that the number of unique templates is significantly lower than the number of logs and the average rate of newly discovered templates decreases as more logs are processed \cite{jiang_lilac_2024}, some LLM-based parsers \cite{jiang_lilac_2024, pei_self-evolutionary_2024} are even able to achieve runtime performance comparable with the fast and well-known parser Drain \cite{he_drain_2017}.

To the best of our knowledge, a structured overview and an objective comparison of papers and methods concerning LLM-based log parsing is currently lacking, as this is a rather novel research field which only recently emerged from the popular research field of generative LLMs. So far there is only a preprint survey paper by Akhtar et al. \cite{akhtar_llm-based_2025}, which provides an overview of using LLMs for log analysis, including log parsing. Although their discussion on log parsing is concise, it lays a valuable foundation for further innovation and development in this area. 
Consequently, our work aims to undertake a systematic review of the literature, identifying the commonalities and variabilities of the existing approaches, defining common terminologies and reporting schemes, highlighting performance enhancing paradigms, and deriving recommendations for future advancements in this field. Our main focus is on investigating LLM-based log parsing methods with emphasis on researchers seeking to apply or develop novel log parsing techniques. Thus, additional emphasis is placed on the application of the LLM and the manual effort required for the implementation of the various approaches.

The aforementioned goals are approached by answering the following research questions:
\begin{itemize}
    \item \textbf{RQ1}: What are the main advantages and disadvantages of LLM-based log parsing approaches and non-LLM-based approaches?
    \item \textbf{RQ2}: To what extent do LLM-based log parsing approaches rely on labeled data and manual configuration?
    \item \textbf{RQ3}: Which techniques can enhance the efficiency or effectiveness of LLM-based log parsing?
    \item \textbf{RQ4}: Which experiment design choices hinder the comparability, and hence which guidelines should be adopted in terms of configuration, datasets used, evaluation metrics, and reporting to make results comparable?
    \item \textbf{RQ5}: To what extent are the presented LLM-based parsing methods accessible and reproducible?
\end{itemize}

We summarize the main contributions of this work as follows:
\begin{itemize}
    \item We provide a systematic overview of the existing methods and techniques based on a set of feature definitions that characterize LLM-based log parsing approaches, or the corresponding papers, respectively.
    \item We derive the general process of LLM-based log parsing, encompassing each parsing approach of each reviewed paper in a single flow chart.
    \item We provide a comprehensive benchmark of seven open-source LLM-based log parsing approaches on open-source datasets.
    \item Based on the literature review and our evaluation results, we answer $5$ research questions, deriving recommendations for future research and the design of (LLM-based) log parsers.
    \item Finally, we make the source code and all the results of our evaluation publicly available in our GitHub repository \cite{beck_github_2025} to ensure transparency and reproducibility.
\end{itemize}

The remainder is structured as follows: Section \ref{sec:background} describes the background and the core concepts. Section \ref{sec:related-work} describes the related work. Section \ref{sec:survey-method} outlines the survey method, including the literature search strategy and the definition of the reviewed features. Section \ref{sec:literature-review-results} presents the results of the literature review in a large table and describes the findings of each feature. Section \ref{sec:evaluation-setup} explains the evaluation setup, while Sec. \ref{sec:evaluation-results} presents the evaluation results. We discuss the findings from literature and evaluation in Sec. \ref{sec:discussion}, and conclude our paper in Sec. \ref{sec:conclusion}.

\section{Background}\label{sec:background}
In this section, we describe the relevant background by means of the core concepts relevant to LLM-based log parsing. 

\subsection{Log Parsing}\label{sec:log-parsing}
Log parsing is the foremost step in log analytics and denotes the extraction of structured data from logs. Logs are given as single- or multi-line events, available in textual form \cite{landauer_deep_2023}. The diverse and un- or semi-structured nature of logs makes log parsing a difficult task. It is the precondition for many log analytics tasks, hence the effectiveness of downstream tasks is inherently dependent on the effectiveness of the parser. Parsing also determines the runtime efficiency of the whole process, which is especially important for online or incremental approaches, such as live-monitoring for cybersecurity \cite{jiang_lilac_2024}. Incremental approaches are designed to handle continuous data streams, updating the parsing model dynamically as new log data arrives, allowing for real-time analysis and quick responses to emerging issues.

In this work, we specifically focus on template extraction, since the extraction of the variables from the log messages is trivial once the (correct) templates are identified, for example, with regular expression matching. The template extraction process is therefore the key component of log parsing.

\subsection{Log Templates}\label{sec:templates}
\begin{figure}[h]
    \centering
    \includegraphics[width=0.8\linewidth]{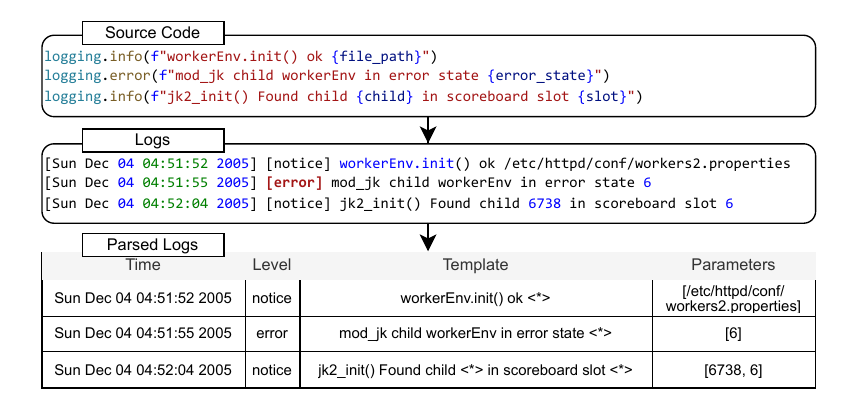}
    \caption{A simple example of log parsing, from logging statement to log file to parsed log.}
    \label{fig:template-example}
\end{figure}

A log template is a predefined format used to log system events in software applications. It typically includes placeholders for essential details like timestamps, log levels (e.g., INFO, ERROR), message content, contextual metadata and more \cite{li_did_2023}. Previous research differentiates between the unstructured log message part of a log and the structured log headers \cite{he_evaluation_2016}. The structured part is thereby straightforward to extract with regular expressions. It has been shown that parsing only the unstructured log message part improves the parsing accuracy for conventional (non-LLM-based) parsers \cite{he_evaluation_2016}. Many methods \cite{he_drain_2017, du_spell_2016, xu_divlog_2024, du_spell_2016}, therefore require the log format ($\neq$ log template) of a dataset as an input parameter. An example of a log template from the Apache dataset \cite{zhu_loghub_2023} is given in Fig. \ref{fig:template-example}. The log format parameter for this example would be ``\texttt{[<Time>] [<Level>] <Content>}'', whereas ``Time'' and ``Level'' refers to log headers and ``Content'' refers to the log message. In the remainder, we understand the ``log format'' as this parameter defining the positional indicators of the log headers and of the content within different log types, or different datasets, respectively.

\subsection{Large Language Models (LLMs)}\label{sec:llm}

Large Language Models (LLMs) have evolved significantly, transitioning from early statistical and rule-based approaches to deep learning architectures powered by Generative Pretrained Transformers (GPTs). These models leverage self-attention mechanisms and large-scale datasets to generate human-like text, enabling advancements in natural language processing (NLP) tasks such as summarization, translation, content generation, and complex problem-solving. A key enhancement in LLMs is instruction tuning which specializes the pretrained LLMs to follow instructions provided in prompts. Instruction tuning enables effective in-context learning (ICL). Instead of modifying model parameters, ICL allows LLMs to learn from prompts that include instructions, examples, and queries, making them more adaptable to specialized applications \cite{zhang_instruction_2024}.

The term LLM is only loosely defined, hence, for this work, we define LLMs as generative pretrained transformers (GPTs) such as ChatGPT, Mistral or Qwen. Models like BERT \cite{devlin_bert_2019} (encoder-only) are thereby excluded, as we are only interested in generative models that excel in both understanding and generating text, with increased generalization capabilities compared to BERT, making them suitable for a broader range of tasks.

\subsection{Log Parsing with LLMs}  

Large-scale model series like ChatGPT and many more have demonstrated potential in this domain. Some approaches fine-tune smaller LLMs \cite{mehrabi_effectiveness_2024} or BERT \cite{le_log_2023-1} to this specific task while reporting superior parsing performance to state-of-the-art approaches. However, fine-tuning LLMs for log parsing is resource-intensive in terms of computational costs, runtime, and labeling examples, making ICL a compelling alternative \cite{xiao_demonstration-free_2024}. Research shows that ICL enhances LLMs' performance in logical reasoning and fact retrieval, suggesting its viability for structured log analysis \cite{brown_language_2020}. However, LLM inference is also costly. The computational demands of LLMs contribute to inference overhead and network latency, but also energy consumption and thus, an increased carbon footprint. Using LLMs for automation therefore necessitates optimizations for practical and energy efficient deployment \cite{jiang_lilac_2024, xiao_demonstration-free_2024}.  Regardless of the efficiency, LLMs are not inherently designed for log parsing, leading to inconsistencies in token classification and errors in log templates. Inaccuracies in the parsed templates can negatively impact downstream tasks like anomaly detection \cite{zhu_tools_2019}. Despite the aforementioned challenges there is a variety of approaches that address these concerns and provide solutions, making LLM-based log parsing a strong alternative to conventional log parsing.

\section{Related Work}\label{sec:related-work}
Existing surveys and benchmarks have significantly contributed to the understanding and evaluation of log parsing. These works are particularly relevant to our work, as they provide a comprehensive overview of the current state of the field, highlighting key methodologies, challenges, and advancements: The work of Zhu et al. \cite{zhu_tools_2019} provides a comprehensive benchmark of log parsing techniques, as well as LogHub \cite{zhu_loghub_2023}, which became the common dataset collection used in many log analysis evaluations. The labels of the LogHub datasets were corrected by Khan et al. \cite{khan_guidelines_2022} to a unified template style. Jiang et al. \cite{jiang_large-scale_2024} conducted a large-scale evaluation on state-of-the-art log parsing techniques while also providing the large-scale dataset collection, LogHub-2.0. It contains several million annotated logs and has been widely adopted for the evaluation of log analysis techniques \cite{jiang_lilac_2024, xiao_demonstration-free_2024, karanjai_logbabylon_2024, huang_lunar_2024, zhong_logparser-llm_2024}. The large number of logs in LogHub-2.0 facilitated comprehensive efficiency evaluations, which are becoming increasingly important due to the large number of logs generated by modern computer systems \cite{mi_toward_2013, zhang_system_2023}. Similarly to our work, Zhang et al. \cite{zhang_system_2023} classify existing parsing approaches by a set of features, such as the parsing mode or the required amount of preprocessing effort. The researchers contemplate the challenges of state-of-the-art log parsing techniques, more specifically, the scarcity of public datasets, limited generalizability caused by intrinsic limitations, a lack of automation, and insufficient efficiency regarding the implementation.

The integration of LLMs into log analysis methods has been explored in the recent survey paper by Akhtar et al. \cite{akhtar_llm-based_2025}. It provides a comprehensive survey of LLM-based techniques for log analysis, highlighting their potential to enhance efficiency and accuracy in processing substantial volumes of log data. The authors discuss various applications, including anomaly detection, fault monitoring, root cause analysis, threat detection, but also provides an overview log parsing. The paper explores the use of fine-tuning, ICL, and retrieval-augmented generation (RAG) \cite{lewis_retrieval-augmented_2020} for performance improvements in log parsing. It also reviews existing research, identifying common challenges such as the high resource consumption of LLMs, the need for frequent updates, and security concerns with closed-source models.

\section{Survey Method}\label{sec:survey-method}
This section describes the search strategy and the taxonomy of LLM-based log parsing.

\subsection{Search Strategy}\label{sec:search-strategy}
This section describes our search strategy and the inclusion and exclusion criteria derived from the core concepts of Sec. \ref{sec:background} and early insights from LLM-based log parsing approaches. For the keyword search, we determined the following keywords:
\begin{enumerate}
    \item \textbf{log(s)} (in title);
    \item \textbf{LLM(s)}, \textbf{large-language-model(s)}, or \textbf{large language model(s)};
    \item \textbf{parsing}, \textbf{parser(s)}, or \textbf{parse}.
\end{enumerate}
The term "log" or "logs" must be included in the publication title, while the remaining keywords are entered into the default search engine of the respective database. This approach should ensure that the search results are focused on publications where log data is the central aspect. The search statistics are given in Table \ref{tab:search-results}.

\begin{table}[h]
    \centering
    \caption{Keyword search results (status 29-01-2025). ``\#R'' and ``\#I'' stand for the number of results and the number of papers that were still included after all exclusion criteria were applied.}
    \resizebox{\columnwidth}{!}{
    \begin{tabular}{|p{3cm}|p{10cm}|c|c|}
    \hline
    \textbf{Database} & \textbf{Search String} & \textbf{\#R} & \textbf{\#I} \\ \hline
    
    IEEE Xplore & 
    \texttt{("Document Title": "log" OR "Document Title": "logs") AND (LLM OR LLMs OR "large language model" OR "large language models") AND (parser OR parsers OR parsing OR parse)} & 23 & 11\\ \hline
    
    Science-Direct & 
    \texttt{title: ("log?") AND title, abstract, keywords: ("LLM" OR "LLMs" OR "large language model" OR "large language models" OR "large-language-model" OR "large-language-models") AND ("parser" OR "parsers" OR "parsing" OR "parse")} & 6 & 0 \\ \hline
    
    ACM Digital Library & 
    \texttt{[[Title: log] OR [Title: logs]] AND [[All: "llm"] OR [All: "llms"] OR [All: "large language model"] OR [All: "large language models"] OR [All: "large-language-model"] OR [All: "large-language-models"]] AND [[All: "parser"] OR [All: "parsers"] OR [All: "parsing"] OR [All: "parse"]]} & 40 & 9 \\ \hline
    
    Google Scholar & 
    \texttt{intitle:"log" OR intitle:"logs" "LLM" OR "LLMs" OR "large-language-model" OR "large-language-models" OR "large language model" OR "large language models" "parsing" OR "parser" OR "parsers" OR "parse"} & 166 & 29 \\ \hline
    
    Web of Science & 
    \texttt{TI=("log" OR "logs") AND TS=("LLM" OR "LLMs" OR "large language model" OR "large language models" OR "large-language-model" OR "large-language-models") AND TS=("parser" OR "parsers" OR "parsing" OR "parse")} & 2 & 2 \\ \hline

    Springer Link &
    \texttt{Title:(log OR logs) AND (LLM OR LLMs OR "large language model" OR "large language models" OR "large-language-model" OR "large-language-models") AND (parser OR parsers OR parsing OR parse)} & 6 & 0 \\ \hline
    \end{tabular}}
    \label{tab:search-results}
\end{table}

The numbers of search results per database sum up to $176$, excluding duplicates found on multiple databases. The results were then filtered by exclusion criteria. A publication is excluded if,
\begin{itemize}
    \item It is not written in English.
    \item It is not accessible in electronic format.
    \item It is a book, technical report, lecture note, presentation, patent, thesis, or dissertation.
    \item A more recent publication is available that presents the same study (by the same authors), ensuring that this SoK focuses on the latest versions of specific approaches.
    \item It only applies a non-LLM-based approach without introducing novelties to LLM-based log parsing.
    \item It does not apply log parsing or LLMs by definitions of Sec. \ref{sec:background}.
\end{itemize}

By applying these exclusion criteria on both abstract and full text, we omitted $147$ of the initial publications found: $1$ publication was omitted because it was not written in English. $5$ were outdated versions of newer papers. $21$ were omitted because they were books, theses, dissertations, or technical reports. $8$ publications were omitted due to being completely unrelated to the topic. $112$ are related but excluded, as they do not directly cover LLM-based log parsing by the definitions of Sec. \ref{sec:background} or focus other log analytics tasks, but employ parsing-free methods or existing conventional parsers, such as Drain \cite{he_drain_2017} or SPELL \cite{du_spell_2016}.

The final selection consists of $29$ papers. One publication is from the year 2023, one from 2025, while the remaining are all from 2024. Since there is a recent hype around the usage of LLMs for log related issues, we include preprint papers to capture the newest research. There are $12$ preprint papers in the final selection ($\sim 41\%$).

\subsection{Feature Definition}\label{sec:feature-definition}
This section describes each feature and its classes of Table \ref{tab:paper-features}, which we report from the analyzed approaches. The features describe either \textbf{general properties (GP)}, \textbf{processing steps (PS)}, or concern the \textbf{reproducibility (R)} of the approaches. Features concerning the general properties of the approaches are typical classifications of machine learning approaches. The features we categorize as processing steps are taken from the work of Jiang et al. \cite{jiang_lilac_2024} and Pei at al. \cite{pei_self-evolutionary_2024}. For reproducibility, we take inspiration from the work of Olszewski et al. \cite{olszewski_get_2023} where they analyzed the reproducibility of machine learning papers of Tier 1 security conferences and defined a set of questions to determine the reproducibility of research findings. During the review process, we found that these features occur with significant frequencies or with significant variations, validating our selection. As illustrated in Table \ref{tab:paper-features}, this is evident.

In Table \ref{tab:paper-features} features are reported with ``\cmark'' (true), an empty cell (false) or a descriptive name or abbreviation. If a paper provides an unclear answer to the feature in question, we write ``?''. For instance, one papers provides a link to its code repository but the repository is empty, thus we write ``?'' for the corresponding feature. If a paper does not include the feature, we leave it blank.

\subsubsection*{GP-1 --- Supervision}
In log parsing, a labeled log is one for which a template, acting as the ground truth for that log, is available. In machine learning approaches are therefore often classified based on the requirement for labels into supervised \cite{wu_log_2024, liu_uniparser_2022} or unsupervised \cite{xiao_demonstration-free_2024, he_drain_2017} approaches, thus we report:
\begin{itemize}
    \item \textit{Supervised parsing (sup)} requires at least some log instances of the training set to be labeled.
    \item \textit{Unsupervised parsing (un)} does not require labels.
\end{itemize}

\subsubsection*{GP-2 --- Parsing Mode}
Many methods are described as online approaches \cite{he_drain_2017, liu_interpretable_2024}, but their interpretations vary. Some consider online processing to be incremental processing or streaming, while others apply batch-wise processing, yet still label their method as online. Additionally, some works do not specify whether their approach is online or offline. Our initial incentive was to classify these methods accordingly, but this is often not feasible due to a lack of context or sufficient explanations. As a consequence, we devised the following categories that describe how many logs are processed at once:
\begin{itemize}
    \item \textit{Stream (S)}: The log lines are processed one-by-one. 
    \item \textit{Batch (B)}: Multiple log lines are processed at once, whereby it is possible to apply local (within each batch) optimizations. A batch is significantly smaller than the entire dataset.
    \item \textit{Total (T)}: The entire dataset is processed at once, whereby it is possible to apply global optimizations to the process.
\end{itemize}

\subsubsection*{GP-3 --- Learning / Prompting}
We identified four different types of learning or prompt engineering techniques. The type of learning can strongly influence how the prompt is designed which is why we report this in a single feature:
\begin{itemize}
    \item \textit{In-context learning (ZS/FS/S)}: ICL \cite{brown_language_2020} leverages the context provided within a single LLM call to generate responses, adapting to specific needs without any updates to the model's parameters. Thereby, we differentiate between a zero-shot setting (ZS), where the model performs a task based solely on an instruction in the prompt, and few-shot ICL (FS), where a small number of demonstrations is provided in the prompt to guide the model’s behavior. Demonstrations can be retrieved randomly from the dataset or with sophisticated strategies. They can also be static (S) which we report separately.
    \item \textit{Chain of Thought (CoT)}: CoT \cite{wei_chain--thought_2022} refers to a series of subsequent LLM calls, where the (complex) task is broken down into (easier) subtasks or ``thoughts'' by the LLM itself and answered in a step-by-step manner, rather than jumping to the solution directly. The process also enhances transparency since intermediate steps can be monitored by users.
    \item \textit{Fine-tuning (FT)}: In addition to ICL and CoT, which are solely modifications of the prompt, fine-tuning \cite{gao_making_2021} modifies the parameters of the LLM, but mostly affects only the last few layers of the neural network, which are the most decisive ones for the outcome.
    \item \textit{Pretraining (PT)}: Pretraining refers to the initial phase of training the model on a large corpus of text data to learn the structure and patterns of language.
\end{itemize}

\subsubsection*{PS-1 --- Manual Configuration}
This is true if at least one manual configuration step is required (e.g., when a parser is applied to an unseen log type). This includes manual definition of input parameters such as regular expressions, log formats or other essential parameters for different datasets, which is often required for many conventional log parsers such as Drain \cite{he_drain_2017}, Brain \cite{yu_brain_2023}, or Spell \cite{du_spell_2016}, and others featured in the LogPAI log parser repository\footnote{\url{https://github.com/logpai/logparser}; accessed 9-December-2024} \cite{zhu_tools_2019}. This feature does not include optional parameters that are generic enough to be left unchanged for a new log type.

\subsubsection*{PS-2 --- Retrieval-Augmented Generation}
Retrieval-augmented generation, or RAG, is a paradigm where the LLM is provided with information retrieval capabilities. In case of log parsing, most approaches utilize a sampling strategy to include either logs or logs and their templates in the prompt. The LLM should then use the provided context to learn the variability and commonality of logs \cite{li_did_2023} or learn parsing directly from log-template pairs. Many approaches create clusters, trees, buckets or other kinds of aggregations from which they sample their demonstrations for ICL and retrieve them based on some kind of similarity measure. We differentiate two cases:
\begin{itemize}
    \item \textit{Random retrieval (R)}: The process is a random retrieval of demonstrations from training data.
    \item \textit{Strategic retrieval (S)}: There is a refined strategy for retrieving demonstrations from specific data storages (e.g., clusters, lists, etc.).
\end{itemize}

\subsubsection*{PS-3 --- Caching}
Some approaches increase their efficiency by storing parsed logs' templates in some kind of data structure and only call the LLM if a subsequent log does not fit any of the existing templates. These data structures can be tree-like structures, similarity-based clusters, or simply lists.

\subsubsection*{PS-4 --- LLM Usage}
The employed LLM is used in at least one of three different ways:
\begin{itemize}
    \item \textit{Direct parsing (dir)}: The LLM receives one or more logs directly in the prompt and is asked for the corresponding template.
    \item \textit{post-processor (post)}: The LLM is used to post-process the log lines, which is mostly merging similar templates or correcting templates based on new information obtained by subsequent log lines in an online approach.
    \item \textit {Preprocessor (pre)}: The LLM functions as a helper in preprocessing, for example, for identifying the timestamp, the log format, or other relevant features.
\end{itemize}

\subsubsection*{PS-5 --- Template Revision}
This feature describes whether the templates are revised in a post-processing step, such as merging similar templates or correcting templates, based on the information obtained from new logs (in an incremental approach). This step can be done with or without the help of LLMs.

\subsubsection*{R-1 --- Evaluation Data}
Since the performance of parsers is strongly dependent on the data, we report the different types of datasets or dataset collections used in the papers:
\begin{itemize}
    \item \textit{LogHub (L)}: The well-known LogHub \cite{zhu_tools_2019, zhu_loghub_2023} repository is widely used for evaluating log analysis methods. It features $16$ annotated datasets from different domains with $2000$ logs each. 
    \item \textit{Corrected LogHub (CL)}: This is a version of LogHub with corrected templates. The original templates have been observed to have inconsistencies due to inconsistent labeling styles \cite{khan_guidelines_2022}.
    \item \textit{LogHub-2.0 (L2)}: LogHub-2.0 \cite{jiang_large-scale_2024} is based on Loghub and features $14$ large-scale datasets from various domains with numbers of logs in the order of $10^4$ to $10^7$.
    \item \textit{Custom (CA/CU)}: If the parser was evaluated with custom data we report whether it is publicly available (CA) or unavailable (CU).
\end{itemize}

\subsubsection*{R-2 --- Evaluation Metrics}\label{sec:eval-metrics-descriptions}
We report the various evaluation metrics employed to assess the effectiveness of log parsing techniques. Each metric provides a unique perspective on the parsing process, focusing on different aspects of effectiveness:

\begin{itemize}
    \item \textit{Group Accuracy (GA)}: A log message is considered correctly parsed if and only if its event template corresponds to the same group of log messages as the ground truth does. GA is then the number of correctly parsed log messages divided by the total number of log messages \cite{zhu_tools_2019}. GA is also known as RandIndex \cite{rand_objective_1971}.
    \item \textit{F1-score of Group Accuracy (FGA)}: $N_p$ is the number of templates that are generated by a log parser, and $N_c$ the number of templates that are correctly parsed by the log parser. $N_g$ is the actual correct number of templates in the ground truth. The precision of the GA (PGA) is then $N_c/N_p$ and the recall of GA (RGA) is $N_c/N_g$. Then, FGA is the harmonic mean of PGA and RGA \cite{jiang_large-scale_2024}.
    \item \textit{Parsing Accuracy (PA)}: A log is considered correctly parsed if and only if all its static text and dynamic variables are correctly identified. PA is then the number of correctly parsed log messages divided by the total number of log messages \cite{dai_logram_2022}. PA is the same as message-level accuracy (MLA) \cite{liu_uniparser_2022, xu_divlog_2024}.
    \item \textit{Edit Distance (ED)}: ED is the minimum number of operations, such as insertions, deletions, or substitutions, needed to convert one string into the other \cite{nedelkoski_self-supervised_2021}. In log parsing, it is used to calculate the minimum number of operations required to convert the parsed template into the ground-truth template \cite{xiao_demonstration-free_2024}.
    \item \textit{Precision Template Accuracy (PTA)}: A template is correctly parsed from log messages if and only if it is identical (token-by-token) to the ground-truth template(s) of the log messages. PTA is the ratio of correctly identified templates to the total number of identified templates \cite{khan_guidelines_2022}. 
    \item \textit{Recall Template Accuracy (RTA)}: Complementary to PTA, RTA is the ratio of correctly identified templates to the total number of ground-truth templates \cite{khan_guidelines_2022}.
    \item \textit{F1-score of Template Accuracy (FTA)}: Is the harmonic mean of PTA and RTA \cite{jiang_large-scale_2024}.
    \item \textit{Other}: If metrics other than the above are used we write ``other''.
\end{itemize}

GA and PA are sensitive to the total number of log messages, which may be problematic since most log datasets contain a large number of logs but a much smaller number of unique templates \cite{jiang_large-scale_2024}. Therefore, Jiang et al. \cite{jiang_large-scale_2024} proposed to use F-metrics since they are insensitive to class imbalance.

\subsubsection*{R-3 --- Used Models}
A variety of different LLMs are used in the analyzed works. We report only the base models, or the name of the model series, of the used LLMs and only the ones used for the task of log parsing.

\subsubsection*{R-4 --- Code Availability}
For a potential user, it might be essential that a parser is already implemented, thus we report the availability of code repositories. This feature is true if a link was provided to the implementation and if the link does not lead to an empty repository or a non-existent page, and false otherwise.

\subsubsection*{R-5 --- Preprint}
This is true if the reviewed paper is a preprint paper. It is possible that the corresponding code repository to the paper, the results, findings, or interpretations are only preliminary, which is why we report this feature.

\section{Literature Review Results}\label{sec:literature-review-results}

The results of our literature review are aggregated in Table \ref{tab:paper-features}. The following subsections discuss the results and findings of the literature review in the context of this table.

\begin{table*}[t]
    \centering
    \caption{Categorization of the selected papers by features.}
    \resizebox{\linewidth}{!}{%
    \begin{tabular}{|p{2.8cm}|c|c|c|c|c|c|c|c|c|p{2.1cm}|p{2.9cm}|c|c|}
\toprule
& \begin{turn}{90}{\textbf{Supervision}}\end{turn} & \begin{turn}{90}{\textbf{Processing mode}}\end{turn} & \begin{turn}{90}{\textbf{Learning / Prompting}}\end{turn} & \begin{turn}{90}{\textbf{Manual config.}}\end{turn} & \begin{turn}{90}{\textbf{RAG}}\end{turn} & \begin{turn}{90}{\textbf{Caching}}\end{turn} & \begin{turn}{90}{\textbf{LLM usage}}\end{turn} & \begin{turn}{90}{\textbf{Template corr.}}\end{turn} & \begin{turn}{90}{\textbf{Datasets}}\end{turn} & \centering \begin{turn}{90}{\textbf{Metrics}}\end{turn} & \centering \begin{turn}{90}{\textbf{Models}}\end{turn} & \begin{turn}{90}{\textbf{Code availability}}\end{turn} & \begin{turn}{90}{\textbf{Preprint}}\end{turn} \\ \hline
& \multicolumn{3}{|c}{\textbf{General Properties}} & \multicolumn{5}{|c}{\textbf{Processing Steps}} & \multicolumn{5}{|c|}{\textbf{Reproducibility}} \\ \hline
\textbf{Approach} & \textbf{GP-1} & \textbf{GP-2} & \textbf{GP-3} & \textbf{PS-1} & \textbf{PS-2} & \textbf{PS-3} & \textbf{PS-4} & \textbf{PS-5} & \textbf{R-1} & \centering \textbf{R-2} & \centering \textbf{R-3} & \textbf{R-4} & \textbf{R-5} \\ \hline
\midrule
Astekin et al. \cite{astekin_comparative_2024} & un, sup & S & ZS, S &  &  &  & dir &  & CL & \centering PA, ED, other & \centering GPT, Claude, Llama &  &  \\ \hline
Astekin et al. \cite{astekin_exploratory_2024} & un, sup & S & S &  &  &  & dir &  & CL & \centering other & \centering GPT, Claude, Llama, Mistral &  &  \\ \hline
Cui et al. \cite{cui_logeval_2024} \mbox{(LogEval)} & un & S & ZS &  & ? &  & dir &  & L & \centering PA, ED & \centering GPT, Claude, Llama, Gemini, InternLM, Qwen, AquilaChat, Mistral, Baichuan, DevOps, ChatGLM & \cmark & \cmark \\ \hline
Fariha et al. \cite{fariha_log_2024}  & un & ? & FS? &  &  &  & pre &  & L & \centering  & \centering GPT &  &  \\ \hline
Huang et al. \cite{huang_lunar_2024} \mbox{(LUNAR)} & un & B & ZS &  & S & \cmark & dir &  & L2 & \centering GA, FGA, PA, FTA & \centering GPT & ? & \cmark \\ \hline
Ji et al. \cite{ji_adapting_2024} \mbox{(SuperLog)} & sup & ? & PT, ZS? &  &  &  & dir &  & L & \centering GA, FTA & \centering Llama, GPT, Qwen, OWL & \cmark & \cmark \\ \hline
Jiang et al. \cite{jiang_lilac_2024} \mbox{(LILAC)} & sup & S & FS & \cmark & S & \cmark & dir & \cmark & L2 & \centering GA, FGA, PA, FTA & \centering GPT& \cmark &  \\ \hline
Karanjai et al. \cite{karanjai_logbabylon_2024} \mbox{(LogBabylon)} & un, sup & S & FT; FS &  & S & \cmark & dir & \cmark & L2 & \centering GA, FGA, PTA, PA, RTA, FTA  & \centering GPT &  & \cmark \\ \hline
Le et al. \cite{le_log_2023} & un, sup & S & ZS, FS &  & R &  & dir &  & CL & \centering GA, PA, ED & \centering GPT &  &  \\ \hline
Liu et al. \cite{liu_interpretable_2024} \mbox{(LogPrompt)} & un & B & ZS & \cmark &  &  & dir &  & L & \centering FTA & \centering GPT, Vicuna & \cmark &  \\ \hline
Ma et al. \cite{ma_llmparser_2024} \mbox{(LLMParser)} & sup & S & FT; ZS, FS & \cmark & ? &  & dir &  & CL & \centering GA, PA & \centering Llama, ChatGLM & \cmark &  \\ \hline
Ma et al. \cite{ma_librelog_2024} \mbox{(OpenLogParser)} & un & B & FS & \cmark & S & \cmark & dir & \cmark & L2 & \centering GA, PA & \centering Llama, Mistral, Gemma, ChatGLM, & \cmark & \cmark \\ \hline
Mehrabi et al. \cite{mehrabi_effectiveness_2024} & sup & S & FT; ZS, S &  &  &  & dir &  & CA & \centering PA, ED, FGA & \centering Mistral, GPT & \cmark &  \\ \hline
Pang et al. \cite{pang_large_2024} (ONLA-LLM) & un, sup & S & FT; ZS &  &  &  & dir &  & CU & \centering PA, PTA, RTA & \centering Llama &  &  \\ \hline
Pei et al. \cite{pei_self-evolutionary_2024} \mbox{(SelfLog)} & un, sup & B & FS & \cmark & S & \cmark & dir & \cmark & L & \centering GA, PA, PTA, RTA & \centering GPT & \cmark &  \\ \hline
Sun et al. \cite{sun_design_2023} (Loki) & un & S & ZS & &  & \cmark & dir &  &  & & &  &  \\ \hline
Sun et al. \cite{sun_semirald_2024} \mbox{(Semirald)} & un, sup & S & FS, CoT & & ? &  & dir &  & L & \centering PA & \centering GPT & \cmark & \cmark \\ \hline
Sun et al. \cite{sun_advancing_2024} \mbox{(SemiSMAC-<T>)} & sup & S & FS, CoT &  & ? & & dir & & L & \centering GA, PA & \centering GPT & ? & \cmark \\ \hline
Vaarandi et al. \cite{vaarandi_using_2024} \mbox{(LLM-TD)} & un & B & S &  &  & \cmark & dir & \cmark & CA & \centering FTA, other & \centering OpenChat, Mistral, WizardLM & \cmark & \cmark \\ \hline
Wu et al. \cite{wu_log_2024} \mbox{(AdaParser)} & un & S & FS & \cmark & S & \cmark & dir, post & \cmark & L, L2 & \centering GA, FGA, PA, FTA & \centering GPT, Gemini, Claude, DeepSeek, Qwen &  & \cmark \\ \hline
Xiao et al. \cite{xiao_demonstration-free_2024} \mbox{(LogBatcher)} & un & B & ZS & \cmark & S & \cmark & dir & \cmark & L2, CL & \centering GA, PA, other  & \centering GPT & \cmark &  \\ \hline
Xu et al. \cite{xu_divlog_2024} \mbox{(DivLog)} & sup & S & FS & \cmark & S &  & dir &  & L & \centering PA, PTA, RTA & \centering GPT & \cmark &  \\ \hline
Xu et al. \cite{xu_help_2024} (HELP) & un, sup & B & ZS, S; CoT & &  & \cmark & dir & \cmark & L2 & \centering GA, FGA, PA, FTA & \centering Claude &  & \cmark \\ \hline
Yu et al. \cite{yu_loggenius_2024} \mbox{(LogGenius)} & un & T? & ZS & \cmark &  &  & pre &  & L, CA & \centering PA & \centering GPT, Gemma & \cmark &  \\ \hline
Zhang et al. \cite{zhang_eclipse_2024} (ECLIPSE) & un & S & ZS & \cmark & ? & \cmark & pre & \cmark & L, CU & \centering GA, FGA & \centering GPT &  & \cmark \\ \hline
Zhang et al. \cite{zhang_lemur_2025} \mbox{(Lemur)} & un & T & FS, CoT & \cmark &  & \cmark & post & \cmark & L & \centering GA, FGA & \centering GPT & \cmark & \cmark \\ \hline
Zhi et al. \cite{zhi_llm-powered_2024} (YALP) & un & S & ZS & \cmark &  & \cmark & dir & \cmark & CL & \centering GA, PA, ED & \centering GPT &  &  \\ \hline
Zhou et al. \cite{zhou_leveraging_2024} & un & S & ZS, FS/S? &  & ? &  & dir &  & L & \centering GA, PA, ED & \centering GPT &  &  \\ \hline
Zhong et al. \cite{zhong_logparser-llm_2024} \mbox{(LogParser-LLM)} & un, sup & S & FT; ZS, FS & & S & \cmark & dir, post & \cmark & L, L2 & \centering GA, PA, FGA, FTA, other & \centering GPT, Llama &  &  \\ \hline
\bottomrule
\end{tabular}
}
    \label{tab:paper-features}
\end{table*}

\subsection{Process Pipeline}

\begin{figure*}[h]
    \centering
    \includegraphics[width=0.85\linewidth]{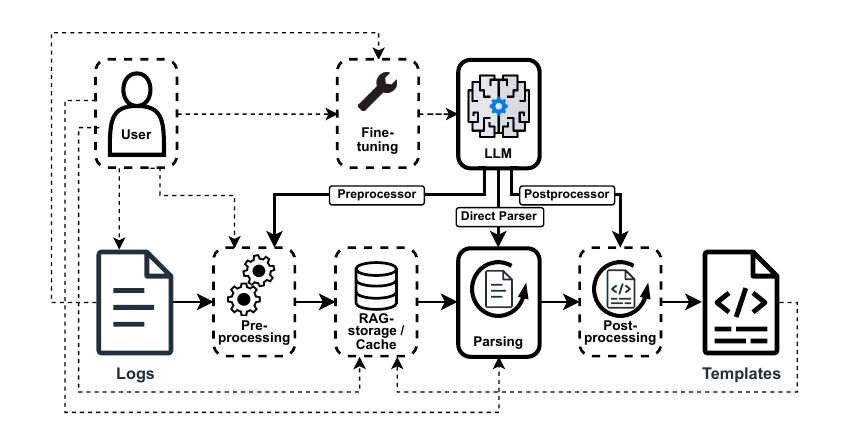}
    \caption{LLM-based parsing process pipeline. The dashed arrows and boxes represent optional components while the continuous ones represent essential components. The arrows pointing from the LLM to certain process elements describe where the LLM can be applied.}
    \label{fig:pipeline}
\end{figure*}

After reviewing the paper selection, we were able to develop a process pipeline that summarizes all the approaches in a single flow chart, given in Fig. \ref{fig:pipeline}. The dashed arrows and boxes represent optional actions or components while the continuous ones represent essential actions components. Optional components can be skipped. The user and the LLM represent key actors that can act on other components. The arrows pointing away from the user indicate steps that can require manual effort. The arrow from user to the logs represent the manual labeling effort for supervised parsers while the arrow pointing to the preprocessing component represents manual configuration (which can also affect other components). We provide detailed explanations about the components in the following sections.

\subsection{Supervision (GP-1)}
From the reviewed papers, $6$ are fully supervised and $9$ approaches describe a supervised and unsupervised setting. Supervised ones require labeled logs for ICL or fine-tuning, but the amount of templates varies from a few guiding seed examples \cite{zhong_logparser-llm_2024} to significant proportions of a dataset's templates \cite{jiang_lilac_2024, xu_divlog_2024}. The works that fine-tune LLMs \cite{ma_llmparser_2024, zhong_logparser-llm_2024, pang_large_2024, karanjai_logbabylon_2024, mehrabi_effectiveness_2024} all require logs and their templates for the process which is why we classify these approaches as supervised.

Similar to earlier work \cite{le_log_2023-1}, LILAC \cite{jiang_lilac_2024} and DivLog \cite{xu_divlog_2024} utilize specialized algorithms to sample from the training data. The objective of the algorithms is to maximize the sample diversity of the labeled logs required for ICL. They call this process candidate sampling. LILAC has a sampling algorithm based on hierarchical clustering while DivLog uses the Determinantal Point Process (DPP) \cite{kulesza_determinantal_2012}. It is noteworthy that the unsupervised parser LogBatcher \cite{xiao_demonstration-free_2024} also uses DPP, but to maximize the sample diversity within the batches of logs that are prompted to the LLM for direct parsing.

Huang et al. (LUNAR) \cite{huang_lunar_2024} empirically study the average FTA of LILAC \cite{jiang_lilac_2024} and the BERT-based parser LogPPT \cite{le_log_2023-1} based on the proportion of labeled logs they receive as input. It was found that FTA exhibited a substantial decline in performance when the proportion of labels was reduced to $5\%$, which, in general, remains a relatively high figure, particularly in light of the typically abundant nature of logs. The finding that the more templates are available, the better the parsers' performance is little surprising \cite{jiang_lilac_2024, huang_lunar_2024}. However, labeled logs are scarce and require substantial manual effort and expertise \cite{xiao_demonstration-free_2024}. All supervised parsers of the selection can technically be operated unsupervised at the cost of performance. This may require modifying the prompt so that it corresponds to a zero-shot setting to not confuse the LLM with announced but absent examples.

\subsection{Processing Mode (GP-2)}
The $18$ parsers classified under the ``stream'' category process logs in real-time as they arrive, making them suitable for online processing. On the other hand, ``batch'' parsers \cite{liu_interpretable_2024, xiao_demonstration-free_2024, xu_help_2024, ma_librelog_2024, pei_self-evolutionary_2024, vaarandi_using_2024} process logs in fixed-sized chunks, meaning they can function in both online and offline settings, depending on the frequency of batch execution. When processing in batches, it is possible to optimize the process for this batch \cite{xiao_demonstration-free_2024}. Lastly, parsers under the category ``total'' \cite{zhang_lemur_2025, yu_loggenius_2024, huang_lunar_2024} analyze an entire log dataset at once, which is inherently an offline approach, as it requires the complete dataset to be available before processing begins. For instance, LUNAR \cite{huang_lunar_2024} and Lemur \cite{zhang_lemur_2025} group the logs based on certain features (e.g., length or most frequent shared token) into buckets. They then create clusters within that buckets that maximize the sample diversity. The concept of maximizing the sample diversity within certain similarity clusters is also used by LogBatcher \cite{xiao_demonstration-free_2024} but for data batches. Similarly, DivLog \cite{xu_divlog_2024} and LILAC \cite{jiang_lilac_2024} use this idea for their sampling algorithm.

\subsection{Learning / Prompting (GP-3)}\label{sec:learning}

\subsubsection{In-Context Learning}
ICL \cite{brown_language_2020} is a prompt-based learning paradigm. It includes learning from instructions but also from examples \cite{brown_language_2020}. This makes it a cheap way of learning. All approaches therefore apply ICL in some way. Among others, $16$ works use zero-shot ICL, $14$ employ few-shot ICL, and
\cite{vaarandi_using_2024, mehrabi_effectiveness_2024, xu_help_2024, astekin_comparative_2024, astekin_exploratory_2024} use static examples. We report static examples separately from the few-shot setting in order to illustrate which approaches use dynamic examples. This in turn implies RAG (PS-2).

\subsubsection{Chain of Thought}
Lemur \cite{zhang_lemur_2025} uses CoT \cite{wei_chain--thought_2022} to revise similar templates, that may belong to the same event, in three dialogue rounds. These steps include revising the structure, the semantics and finally, a decision if the templates should be merged. The other works applying CoT \cite{sun_advancing_2024, sun_semirald_2024, xu_help_2024} use it for direct parsing but do not explain in detail how.

\subsubsection{Fine-Tuning and Pre-Training}
Compared to ICL, fine-tuning is considered a rather computationally costly task \cite{xiao_demonstration-free_2024} since it modifies the parameters of the models itself. Previous research \cite{mosbach_few-shot_2023} found that fine-tuning generally outperforms ICL in effectiveness. From our selection, the works of Ma et al. (LLMParser) \cite{ma_llmparser_2024} and Mehrabi et al.\cite{mehrabi_effectiveness_2024} support this finding. However, fine-tuning also requires labeled logs. As the default setting for the training, \cite{mehrabi_effectiveness_2024} used $80\%$ of the available templates and $15$ logs per template, ONLA-LLM \cite{pang_large_2024} used $80\%$ of the labeled logs of their custom dataset, LLMParser \cite{ma_llmparser_2024} used $50$ labeled logs per dataset, and \cite{zhong_logparser-llm_2024} used $32$ per dataset. Furthermore, Mehrabi et al. \cite{mehrabi_effectiveness_2024} and Ma et al. \cite{ma_llmparser_2024} find that their fine-tuned LLMs struggle with new and unseen log types in terms of effectiveness and robustness. This raises the question of whether this is due to overfitting, but this is a task for future research. It remains up to the users to decide whether a costly training phase, requiring considerable quantities of labeled logs, but with improved performance on seen logs, is feasible for their use case.

From our selection only Ji et al. (SuperLog) \cite{ji_adapting_2024} pretrained a model. Pretraining requires vast amounts of data. To this end, Ji et al. created NLPLog, a comprehensive dataset with a quarter million question-answer pairs on log analysis tasks. The researchers report superior performance not only in parsing but also anomaly detection, failure diagnosis, and log interpretation. They also conducted an evaluation on unseen logs, but they do not report any of the common parsing metrics from Sec. \ref{sec:eval-metrics-descriptions}. This hinders the determination of the practical benefits in settings involving unseen logs. Future research could be conducted on this aspect.

\subsection{Manual Configuration (PS-1)}\label{sec:manual-configuration}
The datasets from LogHub and its descendants \cite{zhu_tools_2019, zhu_loghub_2023, khan_guidelines_2022} have been used for the evaluation of log parsing for many years now and it is known from existing literature which regular expressions, log formats, or other parameters have to be used to achieve a certain baseline of performance. Some of the reviewed methods require multiple parameters that are tuned specifically for each dataset and claim to achieve superior performance on the evaluation metrics. These matters may lead to unrealistic expectations for real-life applications, where sometimes no previous knowledge of the versatile format of logs is given \cite{dai_pilar_2023}. For instance, Dai et al. have shown that the parsing results of (non-LLM-based) parsers like Drain \cite{he_drain_2017}, IPLoM \cite{makanju_clustering_2009}, LenMA \cite{shima_length_2016}, AEL \cite{jiang_abstracting_2008} and Logram \cite{dai_logram_2022} are strongly influenced by their parameter settings. Of the reviewed studies, $12$ require certain configuration parameters. All of them require the log format of the logs to extract the log content from the raw logs and \cite{ma_librelog_2024, zhang_lemur_2025} additionally require regular expressions for preprocessing.

\subsection{RAG (PS-2)}\label{sec:RAG}
Ten approaches describe a sophisticated retrieval process, one describes random retrieval and for six approaches it is not clear how the approaches apply RAG. RAG is used for demonstration retrieval but also for retrieval of multiple target logs where a set of similar logs is chosen. The latter is only relevant for approaches parsing in batches or the whole dataset at once (GP-2). LogBatcher \cite{xiao_demonstration-free_2024}, OpenLogParser \cite{ma_librelog_2024}, and LUNAR \cite{huang_lunar_2024} prompt batches of logs that are in the same similarity cluster but maximize the dissimilarity within that cluster to highlight the variabilities and commonalities in logs for the LLM. Demonstrations can either be only logs \cite{karanjai_logbabylon_2024} or logs and their templates \cite{xu_divlog_2024, jiang_lilac_2024, pei_self-evolutionary_2024, zhong_logparser-llm_2024, astekin_comparative_2024, xiao_demonstration-free_2024, ma_librelog_2024, wu_log_2024}. The templates are either available through being parsed before, yet with no guaranty of correctness, or they origin from candidate sets of labeled logs for supervised settings \cite{xu_divlog_2024,jiang_lilac_2024, sun_advancing_2024, zhong_logparser-llm_2024, pei_self-evolutionary_2024}(GP-1).

As mentioned in \cite{ma_librelog_2024} the quality of demonstrations is of high importance for the parsing effectiveness since it can also introduce noise and confuse the LLM. The works \cite{jiang_lilac_2024, xu_divlog_2024, xiao_demonstration-free_2024, xu_help_2024, zhang_eclipse_2024} support this finding in their ablation studies by comparing the performance of their RAG sampling techniques to random retrieval or static examples, which generally achieves lower scores than strategic retrieval.

Another noteworthy phenomenon observed in generative language models is recency bias, which refers to the tendency of the model to give disproportionate weight or attention to information that appears closer to the end of the prompt \cite{zhao_calibrate_2021}. Consistently, Xu et al. (DivLog) \cite{xu_divlog_2024} observed a significant impact on parsing performance based on the order of the retrieved examples in the prompt, whereby an ascending order of examples by similarity achieved the best results. The ordering of demonstrations in ascending order is adopted by multiple works \cite{xu_divlog_2024, wu_log_2024, xiao_demonstration-free_2024, jiang_lilac_2024}.

\subsection{Caching (PS-3)}\label{sec:caching}
As LLM calls are costly compared to traditional algorithms, many approaches therefore create certain structures like clusters \cite{sun_advancing_2024, xiao_demonstration-free_2024, xu_help_2024, zhang_lemur_2025, jiang_lilac_2024, karanjai_logbabylon_2024, huang_lunar_2024, pei_self-evolutionary_2024}, prefix trees \cite{ma_librelog_2024, jiang_lilac_2024, wu_log_2024, karanjai_logbabylon_2024, zhong_logparser-llm_2024, karanjai_logbabylon_2024} or other structures containing logs and their already parsed templates. In case the target log (or logs) matches an existing template (from an already parsed log) or is at least very similar, the parsing algorithm skips the LLM call and returns the corresponding template retrieved from the structure for this log. This self-evolutionary process \cite{pei_self-evolutionary_2024} is visualized in Fig. \ref{fig:pipeline} by the arrow pointing from the templates to the cache component. This approach can greatly increase efficiency, allowing the usage of parsing with LLMs in settings where fast and cost efficient parsing is crucial, with reported runtimes comparable to runtime efficient parsers, like Drain \cite{he_drain_2017}. A parsing cache can also enhance effectiveness by mitigating the instability of results that often occurs with generative language models \cite{jiang_lilac_2024}. Astekin et al. \cite{astekin_exploratory_2024} find that even with temperature $0$ some models do not answer deterministically. This emphasizes the usefulness of caches even more. However, it is possible that the first retrieved and stored template for an event type is incorrect, even if it matches some of the incoming logs. Such an event can be counteracted by a template revision step (PS-5). All but one paper \cite{sun_design_2023}, which described an approach with some form of caching, also described a post-processing step for correcting templates.

Since the total number of templates is drastically less than the number of logs, LLM parsing becomes a scalable approach with caching \cite{zhong_logparser-llm_2024}. For example, the LogHub-2.0 datasets \cite{jiang_large-scale_2024} contain more than $50$ million logs but less than $3500$ templates. $14$ approaches apply some form of caching, yet some do not explicitly name it ``caching'' and describe it simply as template matching. The structures used for caching can be efficiently and effectively used for sampling similar logs or relevant parsing demonstrations for ICL with RAG. Caching is not bound to the usage with LLMs and can be built into any other parsing approach.

\subsection{LLM Usage (PS-4)}\label{sec:LLM-Usage}
We observed three different ways an LLM can be applied: preprocessing, post-processing and direct parsing, from which we derived the main components of the flow chart in Fig. \ref{fig:pipeline}.

Three approaches employ the LLM for preprocessing. ECLIPSE \cite{zhang_eclipse_2024} applies the LLM for the extraction of semantic information from the logs that is then used by an LCS- (longest common subsequence) and entropy-based parsing algorithm. Fariha et al. \cite{fariha_log_2024} used an LLM for extracting the regular expressions necessary for subsequent parsing with SPELL \cite{du_spell_2016}. LogGenius \cite{yu_loggenius_2024} uses an LLM to augment log groups with low diversity. The actual parsing task is then left to existing unsupervised parsers, like Drain \cite{he_drain_2017} or SPELL \cite{du_spell_2016}, for which their parsing effectiveness should be enhanced by the increased diversity of the logs.

Five approaches perform post-processing with the LLM. In all cases, this is done to correct the already parsed logs to yield more accurate templates. How this is performed is explained in Sec \ref{sec:template-correction}.

Direct parsing is the most straightforward way of parsing with LLMs which is utilized in all but $6$ approaches. There are differences in how the approaches prompt the LLM, yet they all have a similar structure. The prompts these approaches use consist of the following components --- note that the components are not strictly separated from each other and can be interlaced:
\begin{itemize}
    \item \textbf{Instruction}: Instructs the LLM to parse the log(s) and how. The instruction may also contain rules, such as replacement rules for the timestamp or IP address \cite{sun_advancing_2024} or special treatment rules for logs concerning exceptions, errors or similar.
    \item \textbf{Context explanations (optional)}: This part contains explanations about logs or log parsing. For instance, Zhong et al. \cite{zhong_logparser-llm_2024} state to improve the LLM's capabilities of variable identification by including explanations about variable categories as outlined in \cite{li_did_2023}. Other approaches \cite{huang_lunar_2024, sun_advancing_2024, zhong_logparser-llm_2024, sun_semirald_2024, astekin_comparative_2024, xu_help_2024} also provide information about variable types in the prompt. The categorization of the variables can also be beneficial for downstream tasks and the interpretability of the results \cite{li_did_2023, zhong_logparser-llm_2024}.
    \item \textbf{Context examples (optional)}: This part either contains multiple logs to illustrate their variabilities and commonalities or logs and their templates as parsing demonstrations. This part can be static, or dynamic using RAG. Providing examples standardizes the LLM's output format, leading to more stable results \cite{ma_librelog_2024}.
    \item \textbf{Output constraints (optional)}: This part instructs the LLM how to format the output. Some approaches enforce JSON format, but most approaches determine one or two markers (e.g., often backticks or something like /START and /END) so that in a post-processing step the generated template can be extracted easily with regex after the marker or between markers. This improves output quality, because generative language models sometimes generate unwanted output.
    \item \textbf{Target log(s)}: Contains the log(s) to be parsed. The approaches \cite{vaarandi_using_2024, xiao_demonstration-free_2024, huang_lunar_2024, xu_help_2024, ma_librelog_2024, pei_self-evolutionary_2024} provide multiple target logs at once in the query, while the rest provide single logs. Providing multiple logs at once can help the LLM in abstracting variable and static parts from the logs \cite{pei_self-evolutionary_2024, xiao_demonstration-free_2024}.
\end{itemize}

Le et al. \cite{le_log_2023} find that with a simple prompt, where no context is provided nor a detailed instruction is given, GPT-3.5 is hardly able to understand the concept of log parsing. Sun et al. \cite{sun_semirald_2024} achieve a significantly higher PA ($20\%$ to $60\%$) with a prompt that contains extraction rules for direct parsing (on the HDFS and BGL datasets from LogHub \cite{zhu_loghub_2023} with GPT-3.5 and GPT-4) than without extraction rules.

\subsection{Template Revision (PS-5)}\label{sec:template-correction}
In total, $14$ papers revise templates in a post-processing step. All works that correct their templates in a postprocessing step also use a cache (PS-3). The works \cite{zhang_lemur_2025, zhong_logparser-llm_2024, wu_log_2024, liu_interpretable_2024} employ the LLM to revise templates per prompt: Lemur \cite{zhang_lemur_2025} uses CoT in three dialogue rounds for revising semantically similar templates for potential merge. First, the structure is revised, then the semantics, and in a final round the LLM is asked for a solution based on the first two rounds. LogParser-LLM \cite{zhong_logparser-llm_2024} and LogBabylon \cite{karanjai_logbabylon_2024} generate a parsing tree from the templates during parsing. In case, a loose match is identified the LLM decides if the template should be merged, leading to an update of the template tree, or the creation of a new tree root node. The works \cite{liu_interpretable_2024, ma_librelog_2024, wu_log_2024} re-prompt the logs to the LLM, with the same prompt they were initially parsed, to the LLM if they do not match all the related logs until all logs are matched \cite{liu_interpretable_2024} or if a certain number of re-prompts is exceeded \cite{ma_librelog_2024}. AdaParser \cite{wu_log_2024} also revises templates if their variables contain certain keywords linked to exceptions, failures or interrupts, which they state should not be abstracted with a wildcard \cite{wu_log_2024}.

Updating the previously parsed templates can adapt the parser to changes in the computer system and help correct faulty templates. Templates are typically faulty because they have static parts of the log identified as a variable, or variable parts were identified as static. Faulty templates can be identified by high similarity with an existing template \cite{jiang_lilac_2024, zhang_eclipse_2024, zhong_logparser-llm_2024}, if not all logs within a similarity cluster can be matched \cite{xiao_demonstration-free_2024}, or if the centroids of log clusters move closer together through newer incoming logs added to clusters \cite{xu_help_2024}. Templates can be merged with templates that partially match with an existing template, by traversing a prefix tree \cite{jiang_lilac_2024, karanjai_logbabylon_2024, zhong_logparser-llm_2024}, LCS \cite{zhang_eclipse_2024, zhi_llm-powered_2024}, implicitly by reentering the logs into the parsing queue \cite{xiao_demonstration-free_2024}, or by monitoring whether multiple paths (templates) of a prefix tree join together in a subsequent node \cite{pei_self-evolutionary_2024}. A previously parsed template can also be replaced with a new and more permissive template \cite{zhi_llm-powered_2024, vaarandi_using_2024}. More permissive in this context means that more variables were identified for the newer template.

\subsection{Datasets (R-1)}\label{sec:datasets}
All of the papers in our selection use one of the LogHub versions. The most popular dataset is LogHub \cite{zhu_loghub_2023} used in $14$ of the them. LogHub-2.0 \cite{jiang_large-scale_2024} is used $8$ times, and corrected LogHub \cite{khan_guidelines_2022} is used $6$ times. The works \cite{zhang_eclipse_2024, zhong_logparser-llm_2024, xiao_demonstration-free_2024, wu_log_2024, yu_loggenius_2024} evaluate on two or more datasets, or dataset collections. Zhang et al. (ECLIPSE) \cite{zhang_eclipse_2024} and Pang et al. \cite{pang_large_2024} use datasets, other than any of the LogHub ones, that are not publicly available, while \cite{yu_loggenius_2024, mehrabi_effectiveness_2024, vaarandi_using_2024} use custom open-source datasets. The usage of custom datasets along with the datasets from the LogHub versions provides a more comprehensive view on the performance of the parsers. While it is positive to use a variety of datasets for the evaluation, it can also hinder the comparability of the parsing results. For instance, Khan et al. reported performance differences of the conventional parsers on LogHub \cite{zhu_loghub_2023} versus corrected LogHub \cite{khan_guidelines_2022}. Fu et al. \cite{fu_investigating_2022} and Zhang et al. (ECLIPSE) \cite{zhang_eclipse_2024} showed that many state-of-the-art log parsers perform poorly on their custom datasets while performing well on the LogHub datasets \cite{zhu_loghub_2023}.

Using custom datasets alongside LogHub datasets offers a broader view of parser performance but can affect result comparability. The studies \cite{khan_guidelines_2022, fu_investigating_2022, zhang_eclipse_2024} show that parsers often perform well on LogHub datasets \cite{zhu_loghub_2023} or its descendants \cite{jiang_abstracting_2008, khan_guidelines_2022} but poorly on custom datasets. Additionally, it is not clear whether the open-source log datasets used are known to the LLMs from their extensive pretraining phases. However, the ablative studies of \cite{jiang_lilac_2024, xu_divlog_2024, xiao_demonstration-free_2024} report significant performance improvements with ICL compared to direct zero-shot parsing, arguing that this implies a low likelihood of memorization of single log templates for GPT-3 and GPT-3.5-Turbo.

\subsection{Metrics (R-2)}\label{sec:metrics}
Since the correctness evaluation of templates is not straightforward, there is a strong variety of different evaluation metrics. The set of all used metrics is $\{ED, FGA, FTA, GA, PA, PTA, RTA, \texttt{other}\}$. Except for ``other'', these are the traditional metrics that have been widely used \cite{jiang_large-scale_2024, khan_guidelines_2022}. If we compute the Jaccard Index of the used metrics and all available metrics and average the results we get roughly $0.33$, which means that on average only $33\%$ of the full range of metrics is used (including ``other'' as a metric). More specifically, $5$ publications use only PA, or PA and GA, $6$ use ED, $12$ publications use at least one metric that is insensitive to class imbalance (FGA, FTA). Also, $5$ publications use metrics other than traditional ones. For example, the creators of LogBatcher \cite{xiao_demonstration-free_2024} report the normalized Edit Distance (NED) \cite{marzal_computation_1993} which computes the mean ED of all template pairs compared in the data set. In \cite{astekin_exploratory_2024} Astekin et al. report the number of unique templates as a means of LLM determinism, while in \cite{astekin_comparative_2024} they report variations of metics based on ED and LCS. The creators of LogParser-LLM \cite{zhong_logparser-llm_2024} propose metrics based on \textit{Granularity Distance}, the minimum number of operations necessary to transform one parsing result into the other. An operation is thereby defined as converting a static part of the parsed template to a variable or vice versa.

The metrics cover different characteristics of correctness. It is therefore important for evaluating parsers to cover the relevant evaluation aspects with these metrics. For instance, Jiang et al. \cite{jiang_large-scale_2024} proposed using FGA and FTA since they are insensitive to class imbalance, while Khan et al. \cite{khan_guidelines_2022} recommend using GA, PA, RTA and PTA to cover all aspects (FTA is the harmonic mean of RTA and PTA). For example, the anomaly detection tool, DeepLog \cite{du_deeplog_2017}, detects anomalies solely based on the event ID. Each event ID corresponds to a unique template. Consequently, it is indifferent if the parsed templates are correct at the template level as long as they are grouped together correctly, which is indicated by a high score for GA. Since downstream tasks, like anomaly detection, can focus on a variety of data characteristics \cite{beck_semi-supervised_2024}, it is important to report metrics that cover these characteristics.

Previous work \cite{petrescu_log_2023, liu_uniparser_2022, nedelkoski_self-supervised_2021} found that the state-of-the-art log parsers are often evaluated with insufficient selections of evaluation metrics. While their work concerns the conventional non-LLM based approaches, we find that many LLM-based approaches carry on with this manner. More specifically, we consider using only PA or the combination of the grouping metrics GA and FGA as insufficient. The highly varying scores between the metrics of different parsers in \cite{jiang_large-scale_2024, khan_guidelines_2022} but also in our own evaluation in Sec. \ref{sec:evaluation-results} show how important a versatile evaluation metric selection is. Focusing only on a single aspect misleads estimations of a parser's performance.

\subsection{Model (R-3)}
An overview of the LLMs used is given in Table \ref{tab:llms}. We list only the base models or the name of the model series, since there is an overwhelming variety of different versions and sizes of LLMs. The most widely used model series is GPT. The works \cite{ma_llmparser_2024, liu_interpretable_2024, astekin_exploratory_2024, cui_logeval_2024, zhong_logparser-llm_2024, astekin_comparative_2024, ma_librelog_2024, wu_log_2024, yu_loggenius_2024, mehrabi_effectiveness_2024, vaarandi_using_2024, ji_adapting_2024} evaluate multiple LLMs and compare their performance. Since the papers use different evaluation settings (metrics, datasets) or configurations, it is not straightforward to say which LLM performs best. To this end, we refer to the aforementioned works that compare the results of multiple LLMs and specifically the comprehensive benchmark by Cui et al. (LogEval) \cite{cui_logeval_2024} which evaluates $18$ different LLMs (partly from the same series, but with different sizes).
 
\begin{table}[h]
    \centering
    \caption{Overview of LLMs used.}
    \resizebox{0.7\linewidth}{!}{
    \begin{tabular}{lllc}
    \toprule
    \textbf{Base model / Model series}       & \textbf{Creator}           & \textbf{Availability} & \textbf{Times used} \\
    \midrule
    GPT                  & OpenAI                 & API         & 23    \\
    Llama                & Meta                   & Open-source &  8     \\
    Claude               & Anthropic              & API         &  5       \\
    Mistral              & Mistral AI             & Open-source &  5     \\
    ChatGLM              & Tsinghua University    & Open-source &  2     \\
    Gemini               & Google                 & API         &  2     \\
    Gemma                & Google                 & Open-source &  2   \\
    Qwen                 & Alibaba Cloud          & Open-source &  2   \\
    AquilaChat           & BAAI                   & Open-source &  1   \\
    Baichuan             & Baichuan AI            & Open-source &  1     \\
    DeepSeek             & DeepSeek AI            & Open-source &  1   \\
    DevOps-Model         & CodeFuse               & Open-source &  1         \\
    InternLM             & Shanghai AI Lab        & Open-source &  1     \\
    OpenChat             & OpenChat Team          & Open-source &  1     \\
    OWL                  & Camel AI               & Open-source &  1     \\
    Vicuna               & LMSYS                  & Open-source &  1     \\
    WizardLM             & Microsoft              & Open-source &  1     \\
    \bottomrule
    \end{tabular}}
    \label{tab:llms}
\end{table}

\subsection{Code (R-4)}
From our selection, $16$ papers make their implementation available and provide links to their code repositories. At the time of writing, two of them \cite{huang_lunar_2024, sun_advancing_2024} lead to an empty or non-existing repository, but consider that both are preprint versions. The rest did not make their code open-source. A study by Olzewski et al. \cite{olszewski_get_2023} analyzed the reproducibility and replicability of $750$ machine learning papers and their codebases and datasets from Tier 1 security conferences between the years 2013 and 2022. They found that about $59\%$ of the papers did not provide any reproducible artifacts. In our study, $45\%$ of the reviewed papers do not provide any reproducible artifacts (including preprints).

\subsubsection{Code Quality}\label{sec:code-quality}
A large proportion of the approaches do not provide code and therefore can not be used for the evaluation in Sec. \ref{sec:evaluation-results}. Even with available code, some do not provide the comprehensive functionality to replicate the processes described in their papers and comprehensively correcting the code of others is out of scope for this work. At the time of review (February 14th, 2025) the code of Lemur\footnote{\url{https://github.com/zwpride/lemur}; accessed 14-February-2025.} \cite{zhang_lemur_2025} lacks a script for selecting the previously parsed templates for the subsequent CoT template merging process. The code of LogGenius\footnote{\url{https://github.com/huashengyihao/LogGenius}; accessed 14-February-2025.} \cite{yu_loggenius_2024} runs into multiple errors due to missing folders and files, while there is no README file providing instructions. The LogEval repository\footnote{\url{https://github.com/LinDuoming/LogEval}; accessed 14-Febrary-2025.} \cite{cui_logeval_2024} does not provide the scripts for parsing nor instructions. \textbf{LogPrompt}\footnote{\url{https://github.com/lunyiliu/LogPrompt}; accessed 14-February-2025.} \cite{liu_interpretable_2024} encounters multiple errors due to simple typos (e.g. a variable that seems to be undefined but it is actually spelled differently than in its definition). Furthermore, they describe three different prompt strategies, but it is not clear which they used for their result nor do they explain it in the instruction file. Moreover, they do not provide the functionality to run the parsing process except for the self-prompt setting. \textbf{SelfLog} \cite{pei_self-evolutionary_2024} proposes a self-evolutionary approach, in which they ``\textit{retrieve the most similar historical logs and their templates from the data through an Approximate Nearest Neighbors (ANN) search, serving as the corpus for In-Context Learning (ICL)}'' \cite{pei_self-evolutionary_2024}. One could think that this means they update the database incrementally after each newly parsed log. However, in the code\footnote{\url{https://github.com/CSTCloudOps/SelfLog}; accessed 14-February-2025.} the script provided for the effectiveness evaluation does not contain the self-evolution functionality. They provide a second file for online parsing that does provide this functionality but it does not work ad hoc due to missing files and missing function parameters. Also, by default they do not use ANN but cosine similarity for retrieval which is somewhere hidden in the code (not an adjustable parameter). The code repository of \textbf{SuperLog}\footnote{\url{https://github.com/J-York/SuperLog}; accessed 14-Febrary-2025} \cite{ji_adapting_2024} only provides code for LLM training, but not the described framework for parsing. Moreover, we found that many repositories provide sparse instructions, which further impedes reproducibility and application, especially when the execution scripts do not work ad hoc.

These issues appear to be widespread rather than isolated cases. Similar findings were reported in a large-scale study by Trisovic et al. \cite{trisovic_large-scale_2022}, which examined thousands of replication code repositories containing R files published between 2010 and 2020. Their analysis revealed that a significant majority of these files failed to execute properly, even after code cleaning. Likewise, research by Olzewski et al. \cite{olszewski_get_2023} showed that more than half of the artifacts from nearly $300$ reviewed papers could not be run at all. Even among the repositories that did execute, only a fraction produced the expected results, while others either generated different outcomes or lacked crucial components such as arguments or outputs.

\subsubsection{Licenses}
As we focus on an application perspective, we also report the code licenses for the approaches for which the code is available. They are given in Table \ref{tab:approaches-licenses}. Keep in mind that it is possible that code repositories of preprint papers might be preliminary versions or unfinished.
\begin{table}[h]
    \caption{Licenses of the approaches with open-source implementations. Most approaches do not provide a license. Please note, that especially for preprint papers the license might change or, if absent, be added in the future (status 7-March-2025).}
    \centering
    \resizebox{0.8\linewidth}{!}{%
    \begin{tabular}{lcc}
        \toprule
        \textbf{Approach} & \textbf{License} & \textbf{Preprint?} \\
        \midrule
        Cui et al. \cite{cui_logeval_2024} (LogEval) & N/A & \cmark\\
        Ji et al. \cite{ji_adapting_2024} (SuperLog) & Apache License Version 2.0, January 2004 & \cmark \\
        Jiang et al. \cite{jiang_lilac_2024} (LILAC) & N/A & \\
        Liu et al. \cite{liu_interpretable_2024} (LogPrompt) & N/A &\\
        Ma et al. \cite{ma_llmparser_2024} (LLMParser) & N/A &\\
        Ma et al. \cite{ma_librelog_2024} (OpenLogParser) & N/A & \cmark \\
        Mehrabi et al. \cite{mehrabi_effectiveness_2024} & N/A & \\
        Pei et al. \cite{pei_self-evolutionary_2024} (SelfLog) & N/A &\\
        Sun et al. \cite{sun_semirald_2024} (Semirald) & N/A & \cmark\\
        Vaarandi et al. \cite{vaarandi_using_2024} (LLM-TD) & GNU General Public License version 2 & \cmark \\
        Xiao et al. \cite{xiao_demonstration-free_2024} (LogBatcher) & MIT License 2024 & \\
        Xu et al. \cite{xu_divlog_2024} (DivLog) & Apache License Version 2.0, January 2004 & \\
        Yu et al. \cite{yu_loggenius_2024} (LogGenius) & N/A & \\
        Zhang et al. \cite{zhang_lemur_2025} (Lemur) & Apache License Version 2.0, January 2004 & \cmark \\
        \bottomrule
    \end{tabular}}
    \label{tab:approaches-licenses}
\end{table}

\subsection{Notable Mentions}

Besides the features we defined in Sec \ref{sec:feature-definition}, we identified other promising techniques and ideas, worth mentioning in the reviewed papers. For instance, some approaches specifically focus on highlighting the commonalities and variabilities of logs. Lemur \cite{zhang_lemur_2025}, LUNAR \cite{huang_lunar_2024}, OpenLogParser \cite{ma_librelog_2024}, and LogBatcher \cite{xiao_demonstration-free_2024} cluster the logs into multilevel clusters for parsing while the supervised parsers DivLog \cite{xu_divlog_2024} and LILAC \cite{jiang_lilac_2024} use this concept for sampling labeled logs. The coarse-grained clusters thereby capture the commonality of the logs, that, in the best case, all belong to the same log group (i.e. the same event), and the fine-grained clusters should highlight the variabilites within that group. This idea is also used by LogShrink \cite{li_logshrink_2024} for compressing large log files.

Xiao et al. (LogBatcher) \cite{xiao_demonstration-free_2024} report to use the heuristic rules proposed by Khan et al. \cite{khan_guidelines_2022} to correct templates. This includes measures, such as combining subsequent wildcards, like \texttt{<*><*>}, into a single one \texttt{<*>}. This technique, or similar, is also used by \cite{xu_help_2024, jiang_lilac_2024, xu_divlog_2024} and other work outside of our selection \cite{liu_uniparser_2022, le_log_2023-1}. By manually inspecting some the templates generated by the parsers used for the evaluation, we found that a significant proportion of templates would have been considered correct by the evaluation functions, if some of these steps would have been applied.

\section{Evaluation Setup}\label{sec:evaluation-setup}
In terms of evaluation, the existing literature regarding LLM-based log parsing misses a common ground. While there are some commonalities regarding single features, such as used datasets, used evaluation metrics, or used models, the combination of these is highly varying between the approaches - see feature R-1, R-2, and R-3 in Table \ref{tab:paper-features}. As a consequence, we create a benchmark comparing a subset of $7$ out of the $29$ approaches. To counteract randomness from sampling, LLM output and runtime fluctuations, we run each parser $3$ times and take the average.

\subsection{Selection for the Benchmark}
For the benchmark we only select parsers which made their code publicly available --- see feature R-4 in Table \ref{tab:paper-features}. We distinguish between approaches that solely focus on methods changing model parameters, like fine-tuning and pretraining, and approaches that apply the LLM as a tool within a framework. Prior studies showed that fine-tuned \cite{mehrabi_effectiveness_2024} or pretrained \cite{ji_adapting_2024} models can achieve state-of-the-art accuracy scores without pre- and post-processing modules for RAG (PS-2), caching (PS-3), and template revision (PS-5). However, it is clear that improved quality of the LLM's output leads to overall improved output quality of the whole framework. It is therefore meaningful to separate model-centric and model-wrapping approaches --- especially, in the rapidly evolving landscape of LLMs, where more performant models are introduced on a daily basis. For our evaluation, we limit our scope and focus on these frameworks and therefore do not include the fine-tuning or pretraining approaches LLMParser \cite{ma_llmparser_2024}, SuperLog \cite{ji_adapting_2024}, and the work of Mehrabi et al. \cite{mehrabi_effectiveness_2024}, even tough they provide their code or model. We also exclude Semirald \cite{sun_semirald_2024} since it does not provide a log parsing framework and focuses rather on anomaly detection. In Sec. \ref{sec:code-quality}, we described that some approaches provide code, but do not provide functioning parsers. They are consequently not included in the benchmark (except for LogPrompt since its typos could be resolved by exchanging single characters).

Given the above-stated matters the remaining approaches featured in the benchmark are \textbf{LILAC} \cite{jiang_lilac_2024}, \textbf{OpenLogParser} \cite{ma_librelog_2024}, \textbf{LogBatcher} \cite{xiao_demonstration-free_2024}, \textbf{SelfLog} \cite{pei_self-evolutionary_2024}, \textbf{DivLog} \cite{xu_divlog_2024}, \textbf{LLM-TD} \cite{vaarandi_using_2024}, \textbf{LogPrompt} \cite{liu_interpretable_2024}. To a large extent, this selection represents the state-of-the-art of LLM-based log parsing frameworks, as their approaches cover most of the aspects of this field. Unfortunately, due to the aforementioned selection criteria, this benchmark could not include any work using LLMs as pre- or post-processors.

\subsection{Datasets and Baseline}
Following previous work \cite{xiao_demonstration-free_2024, astekin_exploratory_2024, astekin_comparative_2024, zhi_llm-powered_2024, ma_llmparser_2024, le_log_2023, le_log_2023-1} we take the \textbf{corrected version of LogHub} from Khan et al. \cite{khan_guidelines_2022} as the default dataset for our evaluation. It consists of $16$ datasets with ground-truth templates from different domains such as distributed and supercomputer systems, and server application. Note that we found minor errors in the ``Content'' column of the corrected version, mostly additional character spaces, that led to errors when evaluating. We exchanged this column (during runtime) with the ``Content'' column from the original LogHub version \cite{zhu_loghub_2023} (only the templates are different but the content is the same). In Sec. \ref{sec:effectiveness}, we also use the \textbf{original LogHub version} \cite{zhu_loghub_2023} which contains the non-corrected templates. For the runtime evaluation in Sec. \ref{sec:efficiency} we follow previous work \cite{jiang_lilac_2024} and only report the runtime of the large-scale datasets from \textbf{LogHub-2.0} \cite{jiang_large-scale_2024}. This choice is motivated by the need for a more extensive data as the $2000$ samples from the original LogHub are insufficient for a comprehensive runtime analysis. The results for the runtime on the LogHub datasets are available at \cite{beck_github_2025}.

Since the LogHub datasets are so widely used for the evaluation of log parsers (see Sec. \ref{sec:manual-configuration}) there is a tendency that the parsers work especially well on these datasets but not on others. To simulate a use case closer to a real life example, we extend the $16$ datasets by a custom dataset based on AIT Log Dataset V2 \cite{landauer_maintainable_2023, landauer_have_2021}. The dataset contains $2000$ audit logs (from \texttt{russellmitchell/gather/intranet\_server/logs/audit}) with $8$ unique templates. This dataset will be referred to as \textbf{Audit} in the following. The logs were manually annotated by ourselves, adhering to the style of the templates from the corrected LogHub version \cite{khan_guidelines_2022}. Audit logs track system activities, user actions, and changes for security, compliance, and troubleshooting. The corresponding files can be found in our code repository \cite{beck_github_2025}.

For the baseline of the evaluation, we select five non-LLM-based parsers from the LogPAI logparser repository\footnote{\url{https://github.com/logpai/logparser}; accessed 21-March-2025.}, namely \textbf{AEL} \cite{jiang_abstracting_2008}, \textbf{SPELL} \cite{du_spell_2016}, \textbf{Drain} \cite{he_drain_2017}, \textbf{ULP} \cite{sedki_effective_2022} and \textbf{Brain} \cite{yu_brain_2023}. AEL, SPELL and Drain have been widely used in the log parsing research and their performance has been thoroughly investigated by \cite{jiang_large-scale_2024, zhang_system_2023, khan_guidelines_2022}. Their papers were published in 2008 (AEL), 2016 (SPELL) and 2017 (Drain). ULP and Brain are newer advancements from the years 2022 and 2023. The baseline parsers use the standard configuration of the literature \cite{xiao_demonstration-free_2024, jiang_lilac_2024, zhu_loghub_2023, jiang_large-scale_2024}. The default configuration of all these parsers contains the log format, regular expressions (but empty for some datasets), and up to two other parameters that are specific to each of the LogHub datasets. For the Audit dataset \cite{landauer_maintainable_2023, landauer_have_2021} the log format ``\texttt{type=<Type> msg=audit(<Time>): <Content>}'' was used and the regular expression parameter was left empty. The other parameters for the Audit dataset were chosen through grid search by the best score for GA. We used the full dataset for this hyperparameter tuning to show the best possible performance of the baseline.

\subsection{Evaluation Metrics}
Following previous work \cite{jiang_large-scale_2024} we report \textbf{Group Accuracy (GA)}, \textbf{Parsing Accuracy (PA)}, and \textbf{F1-score of Template Accuracy (FTA)} which are described in Sec. \ref{sec:eval-metrics-descriptions}. GA and PA are the most used ones in the literature (cf. Sec. \ref{sec:metrics}) and FTA is recommended by Joang et al. \cite{jiang_large-scale_2024}. Section \ref{sec:eval-metrics-descriptions} also described the Edit Distance (ED), which is the recommended parsing metric of Petrescu et al. \cite{petrescu_log_2023}, from which we compute the \textbf{Normalized Edit Distance (NED)} \cite{marzal_computation_1993}. From our selection, NED is only used by Xiao et al. (LogBatcher) \cite{xiao_demonstration-free_2024}. Nevertheless, we report NED because normalization makes it a more intuitive indicator of effectiveness than ED, as it fits into the $[0,1]$ range of the other conventional metrics.

\subsection{Implementation and Settings}
This section describes the implementation details and the settings used.
\subsubsection{Settings}
The evaluation was performed using three LLMs of different sizes (small, medium, large):
\begin{enumerate}
    \item \textbf{CodeLlama}\footnote{\url{https://ollama.com/library/codellama}; accessed 7-March-2025} (codellama:7b-instruct) is built on top of Llama2 from Meta and has been designed for code-related task. It offers strong performance while requiring significantly fewer parameters than other models. This model is employed to illustrate the performance of a relatively small and efficient LLM ($7$ Billion parameters). Astekin et al. \cite{astekin_comparative_2024} report that CodeLlama outperforms other LLMs (such as GPT-3.5) in parsing accuracy (PA). In a separate study, Astekin et al. \cite{astekin_exploratory_2024} find that both CodeLlama and GPT-3.5 deliver stable results for the log parsing task. The model was run on a Ubuntu 24.04 LTS server with a 15-core Intel Xeon Gold 6226R CPU and a Tesla V100S (32 GB) GPU via Ollama\footnote{\url{https://ollama.com}; accessed 14-March-2025}.
    \item \textbf{GPT-3.5}\footnote{\url{https://platform.openai.com/docs/models/gpt-3.5-turbo}; accessed 7-March-2025} (gpt-3.5-turbo-0125) is a widely known closed-source model, reaching outstanding performance in the earlier stages of the LLM hype. It was chosen because it is the most used one from all reviewed papers and it is considered to be medium sized (the size is actually undisclosed but a paper by researchers of Microsoft \cite{singh_codefusion_2023} state it is 20 billion parameters). The model is proprietary and was accessed via the OpenAI API.
    \item \textbf{DeepSeek R1}\footnote{\url{https://github.com/deepseek-ai/DeepSeek-R1}; accessed 25-February-2025} utilizes a novel reinforcement learning approach for model training. It has recently received wide public attention due to its outstanding performance, while being relatively small (but still large) compared to similarly performing models like OpenAI's o1 model, in combination with its open-source release. This model is used to assess the potential gains in accuracy that can be achieved by utilizing one of the most effective models (671 billion parameters). DeepSeek R1 was called via API from TogetherAI\footnote{\url{https://www.together.ai/}; accessed 14-March-2025}.
\end{enumerate}

The runtime evaluation with the LogHub-2.0 dataset \cite{jiang_large-scale_2024} was run on an Ubuntu 24.04 LTS server with a 32-core AMD EPYC-Milan CPU and 64 GB RAM. With the computation on this hardware and LLM calls to the GPT-3.5 API. In general, we use GPT-3.5 as the default LLM for our evaluation since it is the most used one by the selected approaches.

\subsubsection{Configuration and Code Changes}\label{sec:code-changes}
As mentioned in Sec. \ref{sec:code-quality}, there are some issues with some of the parsers' code which had to be resolved for the benchmark. To ensure the comparability of the output of the parsers, other changes were also necessary to the code of each parser. These changes are kept as small as possible to not disturb the original design or interfere with performance, or at least as little as possible. To ensure model or API compatibility, we updated deprecated versions of the OpenAI (python) package to the newer working ones and we added the functionality to call the OpenAI, Ollama, and TogetherAI API if not given. To have consistent output format, we ensure that the template placeholder symbols are always \texttt{<*>}. Since the parsers LLM-TD \cite{vaarandi_using_2024} and LogPrompt \cite{liu_interpretable_2024} do not only parse the log content but the entire log, we modify the input to these parsers to only parse the content as well. As found by He et al. \cite{he_evaluation_2016}, this typically improves the parsing accuracy.

Specific changes were made to DivLog \cite{xu_divlog_2024}, which is one of the earliest works on LLM-based log parsing. It does not feature a caching mechanism and therefore, calls the LLM for each log line. Since this is a significant burden in terms of runtime (and cost) and therefore also unrealistic for real-life adaptation, we added a simple cache consisting only of the set of already parsed templates. The cache returns the corresponding template in the event of a match instead of calling the LLM. This slightly affects the accuracy metrics due to improved output consistency. Additionally, we implemented a minor post-processing step where multiple consecutive wildcards were replaced by a single wildcard \texttt{<*>} for efficient matching, to resolve an issue with the generation of infinite consecutive wildcards for some of the logs. This measure slightly increases DivLog's accuracy.

LogPrompt \cite{liu_interpretable_2024} accepts a maximum prompt size parameter which is by default $3000$, corresponding to roughly $25$ logs parsed at once (as they state). Early experiments revealed that this performs extremely poor, hence we set the maximum prompt size to $1000$. This corresponds to roughly $5$ to $10$ logs per prompt, which is in the recommended range of simultaneously parsed logs by Xiao et al. \cite{xiao_demonstration-free_2024}. If a template cannot be matched with a log it is again sent to the LLM. We limit the maximum number of re-prompts to $3$ to avoid infinite LLM calls (which occurred in early experiments). LogPrompt also required some code corrections, other than typos. For some datasets, LogPrompt runs into an infinite loop due to a faulty template extraction process from the LLM responses. This was fixed by computing the enumeration of the logs instead of extracting it from the prompt again. From the paper it is not clear whether they used ICL, CoT or the self-prompt paradigm for their parsing performance evaluation, yet the parsing functionality is only given for the self-prompt case.

SelfLog \cite{pei_self-evolutionary_2024} requires the user to set up an SQL database, yet the main script (\texttt{run.py}) does not contain the self-evolution functionality described in the paper (updating the database with new templates). Therefore, we removed the log-groundtruth template examples part from the prompt and operated the parser as unsupervised and without RAG (no database).

\subsubsection{Sampling for Supervised Parsers}
From the selected parsers LILAC \cite{jiang_lilac_2024} and DivLog \cite{xu_divlog_2024} are supervised parsers and require labeled logs to achieve competitive accuracy. In this study, we emphasize minimal human effort and therefore keep the number of available templates $n$ small, namely $n=2$ and $n=4$. Note that the Apache dataset from LogHub \cite{zhu_loghub_2023} contains the smallest number of templates, which is only $6$. For LILAC and DivLog it has been shown that the performance increases with an increasing number of labeled logs available \cite{jiang_lilac_2024, xu_divlog_2024}. This is obvious as the similarity search retrieves the exact template the target log belongs to with a higher likeliness. Multiple works from our selection \cite{jiang_lilac_2024, xu_divlog_2024} as well as \cite{le_log_2023-1} state that having a diverse candidate set is crucial to reduce the risk of overfitting to a specific log template, which is why they create specific sampling algorithms to sample from labeled logs. However, this demands having a large proportion of templates at hand and may require users to label logs manually, thus may hinder real-life application \cite{huang_lunar_2024}.

To ensure comparability, we exchanged the sampling processes of the supervised parsers DivLog \cite{xu_divlog_2024} and LILAC \cite{jiang_lilac_2024} with a simple custom sampling approach: from all labeled logs of a dataset we randomly select $n$ (unique) templates. For each of those templates we randomly select a matching log and thus yield a set of log-template pairs which constitutes the candidate set for the supervised parsers. We sample a new set before each run. In the further course of this paper, we indicate the two supervised parsers by the suffix ``-$n$'', thus LILAC-$n$, DivLog-$n$, with $n\in [2,4]$.

\section{Evaluation Results}\label{sec:evaluation-results}
This section contains the evaluation results with focus on effectiveness and efficiency.

\subsection{Effectiveness}\label{sec:effectiveness}
\subsubsection{Baseline Performance}\label{sec:baseline-performance}
\begin{figure}[h]
    \centering
    \includegraphics[width=0.6\linewidth]{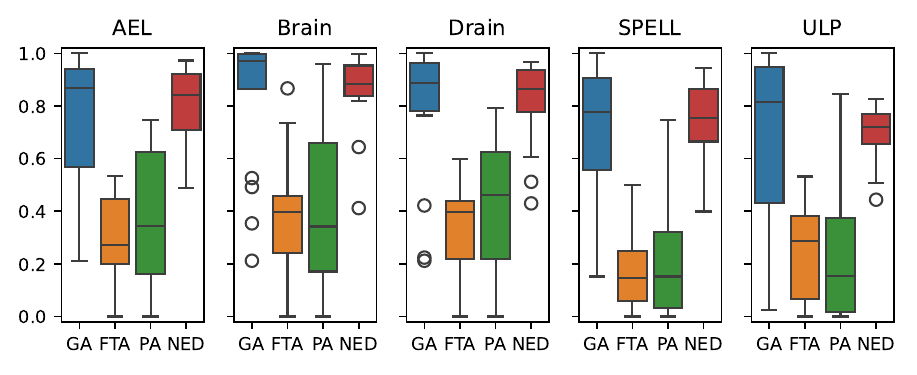}
    \caption{Performance of the baseline parsers on the corrected LogHub dataset including Audit.}
    \label{fig:baseline}
\end{figure}

The performance of the baseline parsers is visualized in Fig. \ref{fig:baseline}. We can see that neither of the baseline parsers achieves satisfactory performance on FTA and PA, meaning that most templates are not matched exactly nor correct when compared token-by-token.

\subsubsection{Performance of LLM-based Parsers}\label{sec:LLM-parsers-performance}
\begin{figure}[h]
    \centering
\begin{subfigure}[b]{\textwidth}
    \centering
    \includegraphics[width=\linewidth]{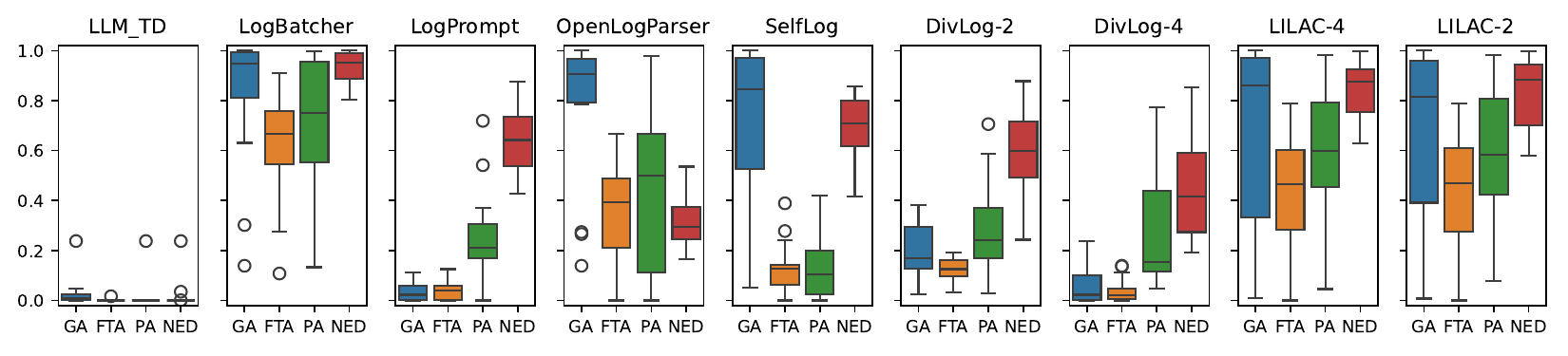}
    \caption{Performance with CodeLlama.}
    \label{fig:codellama}
\end{subfigure}
\begin{subfigure}[b]{\textwidth}
    \centering
    \includegraphics[width=\linewidth]{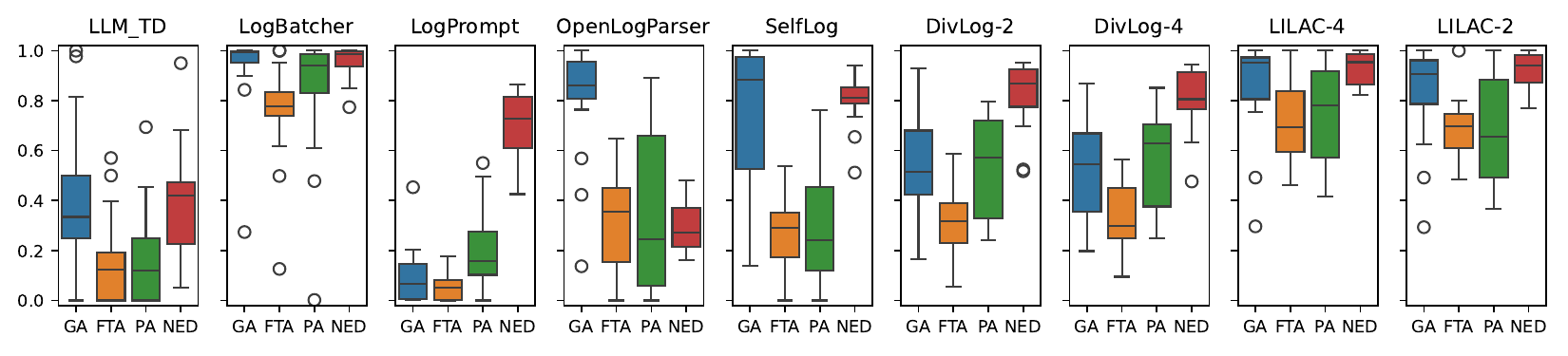}
    \caption{Performance with GPT-3.5.}
    \label{fig:gpt}
\end{subfigure}
\begin{subfigure}[b]{\textwidth}
    \centering
    \includegraphics[width=\linewidth]{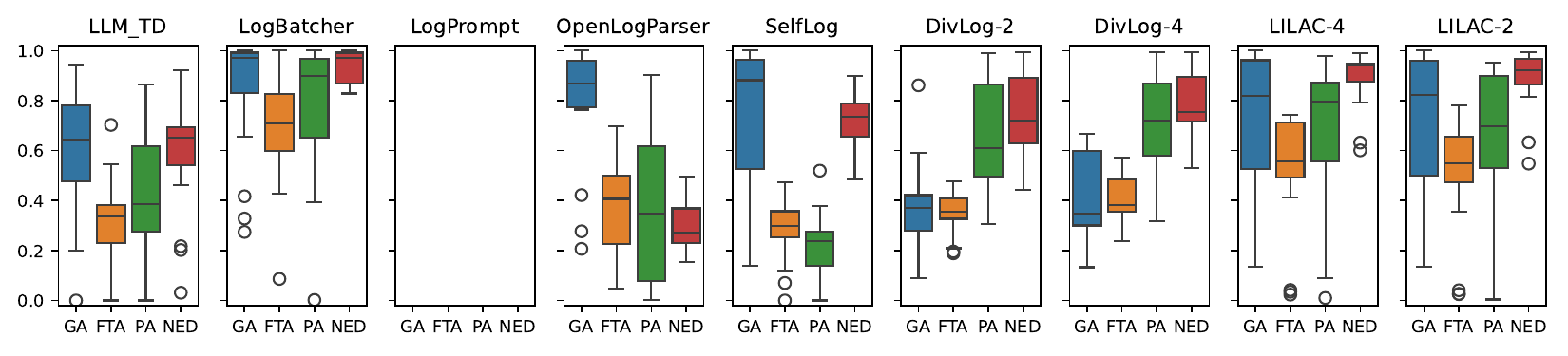}
    \caption{Performance with DeepSeek R1.}
    \label{fig:deepseek}
\end{subfigure}
\caption{Performance of the selected parsers on the corrected LogHub datasets including Audit}
\label{fig:effectiveness}
\end{figure}

The evaluation results of the parsers with the three LLMs on LogHub \cite{zhu_loghub_2023} are given in Fig. \ref{fig:codellama}, Fig. \ref{fig:gpt}, and Fig. \ref{fig:deepseek}. Note that we excluded LogPrompt \cite{liu_interpretable_2024} from the evaluation with DeepSeek R1 because the LLM was barely able to extract any templates, resulting in scores of roughly $0$ in experimental runs. The LLM did not provide the output in the structured format that was requested. Therefore, the template extraction process, which is based on markers positioned directly before the extracted template, was not able to properly extract the template. This resulted in long inner monologues of DeepSeek R1 and maximum reprompts due to log-template mismatches which would have resulted in an unreasonable financial expense given the cost of the LLM and the poor performance.

In the boxplots of Fig. \ref{fig:codellama}, Fig. \ref{fig:gpt}, and Fig. \ref{fig:deepseek} we can see that the best performance on each metric and for each LLM is achieved by LogBatcher \cite{xiao_demonstration-free_2024}, followed by LILAC \cite{jiang_lilac_2024} for $2$ and $4$ shots. LogBatcher and LILAC clearly outperform the conventional parsers, while the conventional parsers outperform the rest of the LLM-based parsers. In general, the performance of each LLM parser is rather stable across the different models. CodeLlama visibly performs worst, but it is also by far the smallest of the models with 7 billion parameters. Given the size difference of GPT-3.5 and DeepSeek R1, it is surprising that DeepSeek R1 does not outperform GPT-3.5. Furthermore, we report the exact numbers for the evaluation with GPT-3.5 in Table \ref{tab:gpt-results} to show how the parsers performed on each individual dataset.

\begin{table}[h]
    \caption{Performance of the selected parsers with GPT-3.5 on the corrected LogHub datasets including Audit. The best scores for each metric on each dataset are marked bold.}
    \centering
    \resizebox{\linewidth}{!}{
\begin{tabular}{cc|ccccccccccccccccc|c}
\toprule
 & Dataset & \begin{turn}{90}{Android}\end{turn} & \begin{turn}{90}{Apache}\end{turn} & \begin{turn}{90}{BGL}\end{turn} & \begin{turn}{90}{HDFS}\end{turn} & \begin{turn}{90}{HPC}\end{turn} & \begin{turn}{90}{Hadoop}\end{turn} & \begin{turn}{90}{HealthApp}\end{turn} & \begin{turn}{90}{Linux}\end{turn} & \begin{turn}{90}{Mac}\end{turn} & \begin{turn}{90}{OpenSSH}\end{turn} & \begin{turn}{90}{OpenStack}\end{turn} & \begin{turn}{90}{Proxifier}\end{turn} & \begin{turn}{90}{Spark}\end{turn} & \begin{turn}{90}{Thunderbird}\end{turn} & \begin{turn}{90}{Windows}\end{turn} & \begin{turn}{90}{Zookeeper}\end{turn} & \begin{turn}{90}{Audit}\end{turn} & \begin{turn}{90}{Average}\end{turn} \\
\midrule
\cline{1-20}
\multicolumn{20}{|c|}{\textbf{Baseline}}\\
\cline{1-20}
\multirow[t]{4}{*}{AEL} & GA & 0.77 & \textbf{1.00} & 0.96 & \textbf{1.00} & 0.90 & 0.87 & 0.57 & 0.40 & 0.76 & 0.54 & 0.25 & 0.97 & 0.90 & 0.94 & 0.69 & 0.92 & 0.21 & 0.74 \\
 & FTA & 0.46 & 0.50 & 0.22 & 0.53 & 0.46 & 0.24 & 0.10 & 0.27 & 0.17 & 0.23 & 0.08 & 0.44 & 0.28 & 0.20 & 0.27 & 0.45 & 0.00 & 0.29 \\
 & PA & 0.39 & 0.69 & 0.34 & 0.62 & 0.68 & 0.51 & 0.16 & 0.17 & 0.17 & 0.25 & 0.02 & 0.67 & 0.38 & 0.04 & 0.15 & 0.75 & 0.00 & 0.35 \\
 & NED & 0.90 & 0.92 & 0.84 & 0.93 & 0.97 & 0.70 & 0.56 & 0.74 & 0.71 & 0.92 & 0.60 & 0.97 & 0.71 & 0.71 & 0.90 & 0.92 & 0.49 & 0.79 \\
\cline{1-20}
\multirow[t]{4}{*}{Brain} & GA & 0.86 & \textbf{1.00} & \textbf{0.99} & \textbf{1.00} & 0.94 & 0.95 & \textbf{1.00} & 0.35 & \textbf{0.94} & \textbf{1.00} & 0.49 & 0.53 & \textbf{1.00} & \textbf{0.97} & \textbf{1.00} & \textbf{0.99} & 0.21 & 0.84 \\
 & FTA & 0.37 & 0.67 & 0.24 & 0.87 & 0.46 & 0.20 & 0.40 & 0.40 & 0.33 & 0.23 & 0.21 & 0.74 & 0.44 & 0.34 & 0.45 & 0.57 & 0.00 & 0.41 \\
 & PA & 0.24 & 0.70 & 0.41 & 0.96 & 0.66 & 0.16 & 0.25 & 0.17 & 0.34 & 0.28 & 0.11 & 0.70 & 0.38 & 0.04 & 0.46 & 0.78 & 0.00 & 0.39 \\
 & NED & 0.84 & 0.92 & 0.84 & \textbf{1.00} & 0.88 & 0.64 & 0.88 & 0.82 & 0.86 & 0.95 & \textbf{0.96} & 0.97 & 0.96 & 0.82 & \textbf{0.92} & 0.94 & 0.41 & 0.86 \\
\cline{1-20}
\multirow[t]{4}{*}{Drain} & GA & 0.83 & \textbf{1.00} & 0.96 & \textbf{1.00} & 0.89 & 0.95 & 0.78 & 0.42 & 0.79 & 0.79 & 0.22 & 0.76 & 0.92 & 0.96 & \textbf{1.00} & 0.97 & 0.21 & 0.79 \\
 & FTA & 0.54 & 0.50 & 0.22 & 0.60 & 0.39 & 0.30 & 0.11 & 0.40 & 0.20 & 0.44 & 0.02 & 0.41 & 0.40 & 0.25 & 0.41 & 0.52 & 0.00 & 0.34 \\
 & PA & 0.55 & 0.69 & 0.34 & 0.63 & 0.65 & 0.51 & 0.23 & 0.18 & 0.22 & 0.51 & 0.02 & 0.68 & 0.38 & 0.05 & 0.46 & 0.79 & 0.00 & 0.41 \\
 & NED & 0.90 & 0.92 & 0.84 & 0.93 & 0.96 & 0.79 & 0.61 & 0.78 & 0.76 & 0.97 & 0.51 & 0.96 & 0.96 & 0.81 & 0.86 & 0.94 & 0.43 & 0.82 \\
\cline{1-20}
\multirow[t]{4}{*}{SPELL} & GA & 0.86 & \textbf{1.00} & 0.79 & \textbf{1.00} & 0.65 & 0.78 & 0.64 & 0.15 & 0.76 & 0.56 & 0.26 & 0.53 & 0.90 & 0.84 & 0.99 & 0.96 & 0.34 & 0.71 \\
 & FTA & 0.25 & 0.50 & 0.06 & 0.43 & 0.39 & 0.16 & 0.10 & 0.14 & 0.05 & 0.23 & 0.00 & 0.04 & 0.19 & 0.15 & 0.10 & 0.37 & 0.00 & 0.19 \\
 & PA & 0.15 & 0.69 & 0.20 & 0.30 & 0.53 & 0.11 & 0.15 & 0.09 & 0.03 & 0.19 & 0.00 & 0.48 & 0.32 & 0.03 & 0.00 & 0.75 & 0.00 & 0.24 \\
 & NED & 0.81 & 0.92 & 0.67 & 0.94 & 0.84 & 0.49 & 0.52 & 0.75 & 0.66 & 0.86 & 0.50 & 0.93 & 0.67 & 0.71 & 0.77 & 0.94 & 0.40 & 0.73 \\
\cline{1-20}
\multirow[t]{4}{*}{ULP} & GA & 0.74 & \textbf{1.00} & 0.93 & \textbf{1.00} & \textbf{0.95} & \textbf{0.99} & 0.90 & 0.18 & 0.81 & 0.43 & 0.47 & 0.02 & 0.92 & 0.68 & 0.41 & \textbf{0.99} & 0.34 & 0.69 \\
 & FTA & 0.29 & 0.00 & 0.30 & 0.00 & 0.53 & 0.20 & 0.33 & 0.40 & 0.28 & 0.07 & 0.04 & 0.38 & 0.20 & 0.44 & 0.38 & 0.38 & 0.00 & 0.25 \\
 & PA & 0.23 & 0.00 & 0.38 & 0.00 & 0.85 & 0.16 & 0.07 & 0.20 & 0.24 & 0.15 & 0.02 & 0.51 & 0.02 & 0.07 & 0.39 & 0.50 & 0.00 & 0.22 \\
 & NED & 0.67 & 0.74 & 0.69 & 0.61 & 0.77 & 0.72 & 0.80 & 0.61 & 0.69 & 0.78 & 0.66 & 0.78 & 0.77 & 0.74 & 0.44 & 0.82 & 0.51 & 0.69 \\
\cline{1-20}
\multicolumn{20}{|c|}{\textbf{Unsupervised}}\\
\cline{1-20}
\multirow[t]{4}{*}{LLM\_TD} & GA & 0.00 & \textbf{1.00} & 0.50 & 0.98 & 0.22 & 0.47 & 0.24 & 0.39 & 0.00 & 0.25 & 0.82 & 0.27 & 0.29 & 0.26 & 0.57 & 0.35 & 0.33 & 0.41 \\
 & FTA & 0.00 & 0.50 & 0.20 & 0.00 & 0.19 & 0.17 & 0.10 & 0.14 & 0.00 & 0.40 & 0.57 & 0.00 & 0.12 & 0.04 & 0.10 & 0.16 & 0.00 & 0.16 \\
 & PA & 0.00 & 0.69 & 0.45 & 0.00 & 0.20 & 0.12 & 0.24 & 0.28 & 0.00 & 0.36 & 0.25 & 0.00 & 0.23 & 0.10 & 0.01 & 0.11 & 0.00 & 0.18 \\
 & NED & 0.05 & 0.95 & 0.51 & 0.43 & 0.26 & 0.47 & 0.28 & 0.42 & 0.06 & 0.68 & 0.46 & 0.22 & 0.42 & 0.29 & 0.48 & 0.18 & 0.09 & 0.37 \\
\cline{1-20}
\multirow[t]{4}{*}{LogBatcher} & GA & \textbf{0.97} & \textbf{1.00} & \textbf{0.99} & \textbf{1.00} & \textbf{0.95} & \textbf{0.99} & \textbf{1.00} & \textbf{0.84} & 0.92 & \textbf{1.00} & \textbf{0.98} & \textbf{1.00} & \textbf{1.00} & 0.90 & \textbf{1.00} & \textbf{0.99} & 0.27 & \textbf{0.93} \\
 & FTA & \textbf{0.78} & \textbf{1.00} & 0.83 & \textbf{1.00} & \textbf{0.75} & \textbf{0.74} & \textbf{0.95} & \textbf{0.75} & \textbf{0.50} & 0.82 & 0.79 & \textbf{1.00} & \textbf{0.76} & \textbf{0.67} & \textbf{0.62} & \textbf{0.81} & 0.13 & \textbf{0.76} \\
 & PA & \textbf{0.83} & \textbf{1.00} & 0.94 & \textbf{1.00} & \textbf{0.94} & \textbf{0.89} & \textbf{0.99} & \textbf{0.84} & 0.48 & \textbf{0.97} & \textbf{0.77} & \textbf{1.00} & \textbf{0.97} & \textbf{0.84} & 0.61 & \textbf{0.99} & 0.00 & \textbf{0.83} \\
 & NED & \textbf{0.92} & \textbf{1.00} & \textbf{0.99} & \textbf{1.00} & \textbf{1.00} & \textbf{0.97} & \textbf{1.00} & \textbf{0.94} & 0.86 & \textbf{0.99} & 0.94 & \textbf{1.00} & 0.98 & \textbf{0.96} & 0.85 & \textbf{1.00} & 0.77 & \textbf{0.95} \\
\cline{1-20}
\multirow[t]{4}{*}{LogPrompt} & GA & 0.13 & 0.19 & 0.14 & 0.00 & 0.17 & 0.07 & 0.06 & 0.12 & 0.20 & 0.00 & 0.01 & 0.00 & 0.01 & 0.07 & 0.12 & 0.45 & 0.00 & 0.10 \\
 & FTA & 0.12 & 0.00 & 0.03 & 0.00 & 0.07 & 0.08 & 0.06 & 0.15 & 0.09 & 0.00 & 0.00 & 0.00 & 0.03 & 0.18 & 0.05 & 0.07 & 0.00 & 0.06 \\
 & PA & 0.24 & 0.16 & 0.38 & 0.00 & 0.55 & 0.15 & 0.27 & 0.13 & 0.23 & 0.13 & 0.10 & 0.00 & 0.25 & 0.09 & 0.33 & 0.50 & 0.00 & 0.21 \\
 & NED & 0.78 & 0.85 & 0.83 & 0.49 & 0.85 & 0.70 & 0.71 & 0.61 & 0.77 & 0.73 & 0.43 & 0.54 & 0.82 & 0.69 & 0.78 & 0.86 & 0.54 & 0.70 \\
\cline{1-20}
\multirow[t]{4}{*}{OpenLogParser} & GA & 0.85 & \textbf{1.00} & 0.96 & 0.85 & 0.93 & 0.96 & 0.82 & 0.42 & 0.81 & 0.86 & 0.57 & 0.76 & 0.87 & 0.94 & 0.99 & 0.96 & 0.14 & 0.81 \\
 & FTA & 0.32 & 0.50 & 0.42 & 0.00 & 0.59 & 0.28 & 0.46 & 0.45 & 0.27 & 0.65 & 0.05 & 0.00 & 0.41 & 0.36 & 0.15 & 0.41 & 0.00 & 0.31 \\
 & PA & 0.24 & 0.69 & 0.72 & 0.00 & 0.89 & 0.09 & 0.66 & 0.24 & 0.26 & 0.69 & 0.06 & 0.00 & 0.36 & 0.06 & 0.01 & 0.48 & 0.00 & 0.32 \\
 & NED & 0.17 & 0.46 & 0.37 & 0.27 & 0.30 & 0.21 & 0.26 & 0.27 & 0.16 & 0.39 & 0.24 & 0.37 & 0.21 & 0.19 & 0.44 & 0.27 & 0.48 & 0.30 \\
\cline{1-20}
\multirow[t]{4}{*}{SelfLog} & GA & 0.89 & \textbf{1.00} & 0.97 & \textbf{1.00} & 0.89 & \textbf{0.99} & 0.92 & 0.29 & 0.79 & 0.44 & 0.44 & 0.53 & 0.94 & 0.70 & 0.72 & \textbf{0.99} & 0.14 & 0.74 \\
 & FTA & 0.39 & 0.33 & 0.29 & 0.00 & 0.16 & 0.27 & 0.48 & 0.31 & 0.24 & 0.17 & 0.54 & 0.00 & 0.38 & 0.35 & 0.31 & 0.24 & 0.00 & 0.26 \\
 & PA & 0.46 & 0.28 & 0.66 & 0.00 & 0.22 & 0.14 & 0.55 & 0.09 & 0.29 & 0.15 & 0.31 & 0.12 & 0.76 & 0.08 & 0.59 & 0.24 & 0.00 & 0.29 \\
 & NED & 0.76 & 0.85 & 0.94 & 0.86 & 0.65 & 0.82 & 0.89 & 0.79 & 0.80 & 0.85 & 0.73 & 0.79 & 0.93 & 0.82 & 0.81 & 0.80 & 0.51 & 0.80 \\
 \cline{1-20}
\multicolumn{20}{|c|}{\textbf{Supervised}}\\
\cline{1-20}
\multirow[t]{4}{*}{DivLog-2} & GA & 0.56 & 0.91 & 0.60 & 0.49 & 0.52 & 0.84 & 0.68 & 0.16 & 0.38 & 0.47 & 0.30 & 0.72 & 0.46 & 0.36 & 0.52 & 0.93 & 0.42 & 0.55 \\
 & FTA & 0.39 & 0.48 & 0.32 & 0.06 & 0.23 & 0.31 & 0.57 & 0.32 & 0.19 & 0.33 & 0.11 & 0.39 & 0.23 & 0.31 & 0.19 & 0.59 & 0.48 & 0.32 \\
 & PA & 0.52 & 0.70 & 0.72 & 0.57 & 0.77 & 0.32 & 0.76 & 0.33 & 0.26 & 0.64 & 0.24 & 0.76 & 0.67 & 0.35 & 0.31 & 0.80 & 0.49 & 0.54 \\
 & NED & 0.78 & 0.92 & 0.93 & 0.76 & 0.90 & 0.80 & 0.92 & 0.79 & 0.52 & 0.95 & 0.52 & 0.95 & 0.87 & 0.82 & 0.70 & 0.92 & 0.94 & 0.82 \\
\cline{1-20}
\multirow[t]{4}{*}{DivLog-4} & GA & 0.67 & 0.53 & 0.55 & 0.65 & 0.36 & 0.70 & 0.86 & 0.20 & 0.59 & 0.59 & 0.27 & 0.32 & 0.67 & 0.40 & 0.52 & 0.87 & 0.36 & 0.53 \\
 & FTA & 0.42 & 0.43 & 0.22 & 0.25 & 0.18 & 0.25 & 0.56 & 0.34 & 0.30 & 0.55 & 0.10 & 0.29 & 0.48 & 0.30 & 0.22 & 0.53 & 0.45 & 0.34 \\
 & PA & 0.50 & 0.70 & 0.67 & 0.71 & 0.53 & 0.32 & 0.76 & 0.38 & 0.36 & 0.65 & 0.25 & 0.85 & 0.77 & 0.27 & 0.63 & 0.79 & 0.56 & 0.57 \\
 & NED & 0.78 & 0.91 & 0.93 & 0.87 & 0.63 & 0.69 & 0.91 & 0.81 & 0.78 & 0.95 & 0.48 & 0.91 & 0.93 & 0.64 & 0.80 & 0.93 & 0.77 & 0.81 \\
\cline{1-20}
\multirow[t]{4}{*}{LILAC-4} & GA & 0.95 & \textbf{1.00} & 0.98 & 0.94 & \textbf{0.95} & 0.97 & 0.91 & 0.30 & 0.81 & 0.75 & 0.49 & 0.95 & 0.97 & 0.94 & 0.79 & \textbf{0.99} & \textbf{1.00} & 0.86 \\
 & FTA & 0.66 & \textbf{1.00} & \textbf{0.89} & 0.46 & \textbf{0.75} & 0.59 & 0.80 & 0.70 & 0.47 & \textbf{0.84} & \textbf{0.83} & 0.92 & 0.67 & 0.56 & 0.57 & 0.69 & \textbf{0.88} & 0.72 \\
 & PA & 0.57 & \textbf{1.00} & \textbf{0.96} & 0.53 & 0.92 & 0.83 & 0.89 & 0.42 & \textbf{0.49} & 0.78 & 0.43 & 0.99 & 0.93 & 0.73 & \textbf{0.70} & 0.64 & \textbf{0.78} & 0.74 \\
 & NED & 0.85 & \textbf{1.00} & \textbf{0.99} & 0.82 & 0.99 & 0.95 & 0.96 & 0.92 & 0.86 & 0.97 & 0.90 & \textbf{1.00} & 0.97 & 0.93 & 0.85 & 0.82 & \textbf{0.99} & 0.93 \\
\cline{1-20}
\multirow[t]{4}{*}{LILAC-2} & GA & 0.96 & \textbf{1.00} & 0.96 & 0.94 & 0.88 & 0.89 & 0.91 & 0.29 & 0.79 & 0.63 & 0.49 & 0.85 & 0.92 & 0.96 & 0.69 & \textbf{0.99} & \textbf{1.00} & 0.83 \\
 & FTA & 0.70 & \textbf{1.00} & 0.80 & 0.50 & 0.70 & 0.57 & 0.80 & 0.61 & 0.48 & 0.75 & 0.74 & 0.77 & 0.66 & 0.61 & 0.57 & 0.70 & 0.67 & 0.68 \\
 & PA & 0.57 & \textbf{1.00} & 0.94 & 0.48 & 0.87 & 0.70 & 0.89 & 0.37 & 0.48 & 0.64 & 0.37 & 0.92 & 0.88 & 0.78 & 0.52 & 0.66 & 0.49 & 0.68 \\
 & NED & 0.87 & \textbf{1.00} & 0.98 & 0.91 & 0.98 & 0.90 & 0.96 & 0.87 & \textbf{0.87} & 0.94 & 0.90 & 0.99 & \textbf{0.99} & 0.95 & 0.77 & 0.87 & 0.98 & 0.93 \\
\cline{1-20}
\bottomrule
\end{tabular}}
    \label{tab:gpt-results}
\end{table}

\subsubsection{LLM General Performance}\label{sec:LLM-performance}
\begin{figure}[h]
    \centering
    \includegraphics[width=0.4\linewidth]{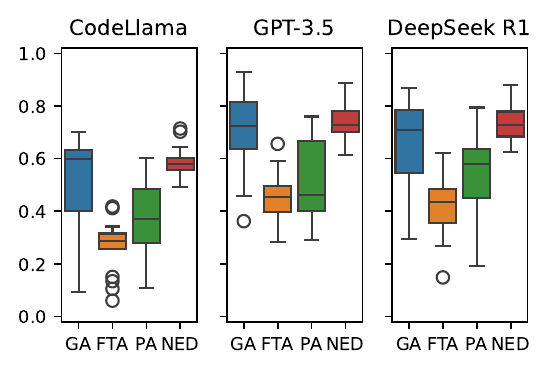}
    \caption{Averaged performance of LLM-based parsers for CodeLlama, GPT-3.5 and DeepSeek R1.}
    \label{fig:models}
\end{figure}

Figure \ref{fig:models} shows the performance for CodeLlama, GPT-3.5 and DeepSeek R1 averaged over all parsers and datasets. One can see that CodeLlama performs worst while GPT-3.5 slightly outperforms DeepSeek R1 except for PA. Consequently, the financial implications of employing DeepSeek R1 do not justify the cost. For instance, the OpenAI API for GPT-3.5 (gpt-3.5-0125) costs $0.5\$$ per $1$ million tokens (input and output) while DeepSeek R1 from the TogetherAI API costs $3\$$ for input and $7\$$ for the output per $1$ million tokens. We found that the reasoning steps of DeepSeek R1 are not always helpful and especially when a structured output is required may confuse the LLM. A possible reason could be that the process of log parsing, not being complex enough, leads to overthinking \cite{cuadron_danger_2025}. However, in \cite{cuadron_danger_2025} the researchers state that DeepSeek R1 is robust against overthinking. The usefulness of reasoning models for log parsing should therefore be investigated in future research.

\subsubsection{Performance Difference on LogHub and corrected LogHub}\label{sec:loghub-vs-corrected-loghub}
\begin{figure}[h]
    \centering
    \includegraphics[width=\linewidth]{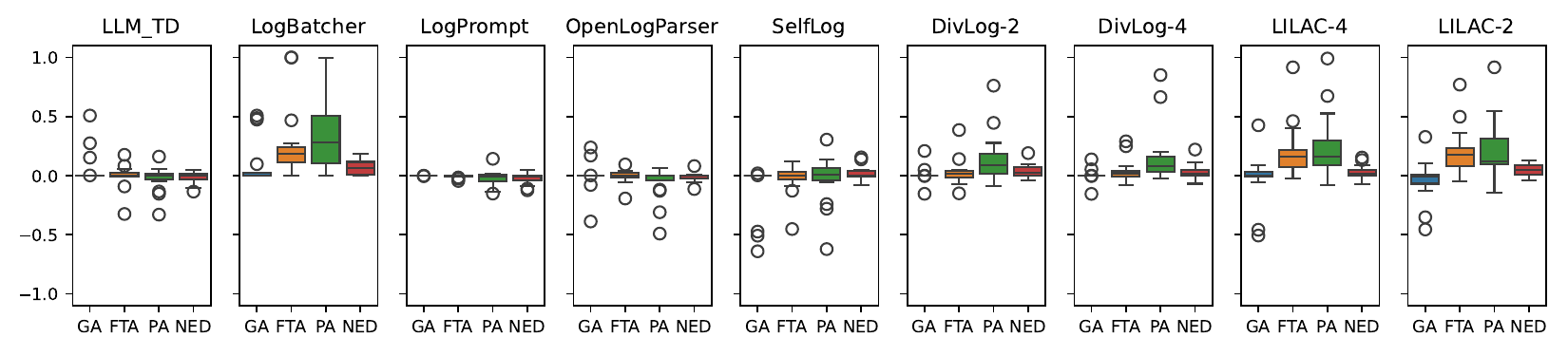}
    \caption{Performance of the parsers with GPT-3.5 on corrected LogHub minus the performance on the original LogHub.}
    \label{fig:difference}
\end{figure}

As in the study by Khan et al. \cite{khan_guidelines_2022}, we report the performance difference between the evaluation on the corrected LogHub datasets \cite{khan_guidelines_2022} and the original LogHub datasets \cite{zhu_loghub_2023} (performance of corrected minus performance of original) with GPT-3.5. In Fig. \ref{fig:difference} we can see that the highest differences are given for LogBatcher \cite{xiao_demonstration-free_2024}, followed by LILAC \cite{jiang_lilac_2024} for both sample numbers. Naturally, LILAC's and DivLog's \cite{xu_divlog_2024} samples were collected and evaluated with the same dataset. That the difference is rather on the positive side indicates that they achieve better performance on the corrected LogHub dataset. Interestingly LILAC was originally evaluated with LogHub-2.0 whose templates correspond to the original LogHub templates and not to the corrected ones \cite{jiang_lilac_2024}. In general, we can see a slight tendency that the parsers output templates that correspond more to the corrected datasets' templates, which suggests that LLMs ``intuitively'' prefer the corrected templates' format.

\subsubsection{Performance on the Audit Dataset}\label{sec:audit-evaluation}

An analysis of the performance of all parsers on the Audit dataset reveals that only the supervised parser, LILAC \cite{jiang_lilac_2024}, and to some extent also DivLog \cite{xu_divlog_2024}, are able to parse the logs effectively. Despite the fact that the parameters of the baseline parsers were hyperparameter tuned, they did not achieve satisfactory results. It appears that the intended templates' format of the logs is not straightforward for conventional parsers, but neither for GPT-3.5 without supervised demonstrations. This finding underscores the efficacy of supervised parsers, demonstrating that a mere two templates are sufficient to attain a reasonable performance level, potentially even a single one demonstrating the preferred template style.

\begin{figure}[h]
    \centering
    \includegraphics[width=0.9\linewidth]{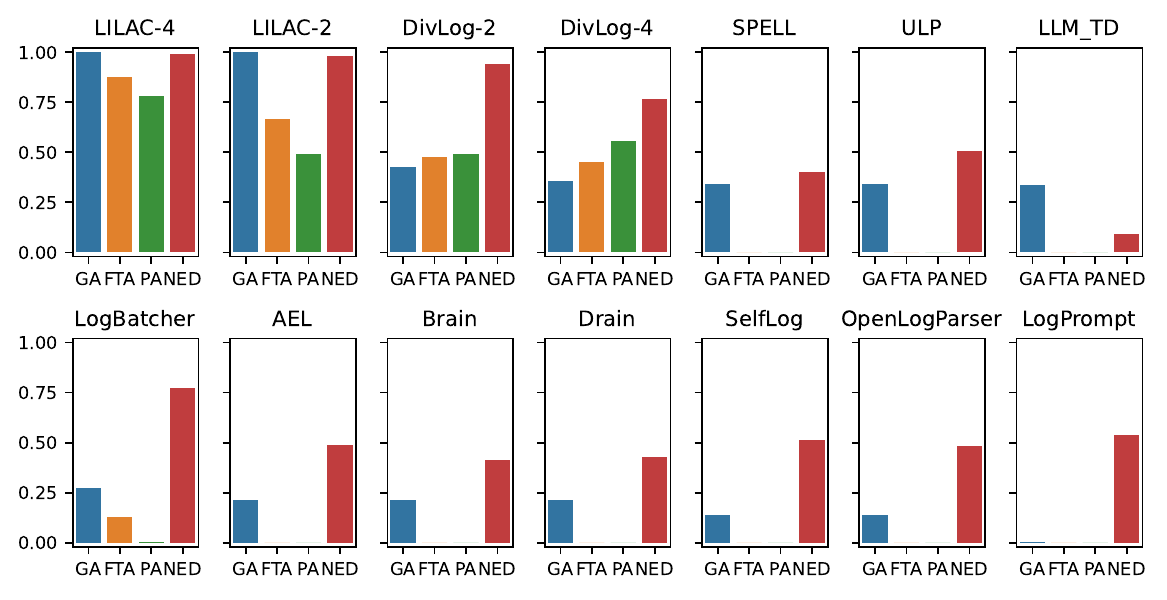}
    \caption{Performance of the selected LLM-based parsers with GPT-3.5 and the baseline on the Audit data.}
    \label{fig:gpt-audit}
\end{figure}

\subsection{Efficiency}\label{sec:efficiency}

For the evaluation regarding efficiency we select the well-known datasets HDFS and BGL from LogHub-2.0 \cite{jiang_large-scale_2024}. HDFS contains $11.2$ million logs with $46$ unique templates while BGL contains $4.6$ million logs with $320$ unique templates. We excluded DivLog \cite{xu_divlog_2024} and LogPrompt \cite{liu_interpretable_2024} from this evaluation since they do not utilize caching. The computation would therefore scale linearly with the number of logs and require multiple million LLM calls.

\begin{figure}[h]
    \centering
    \includegraphics[width=\linewidth]{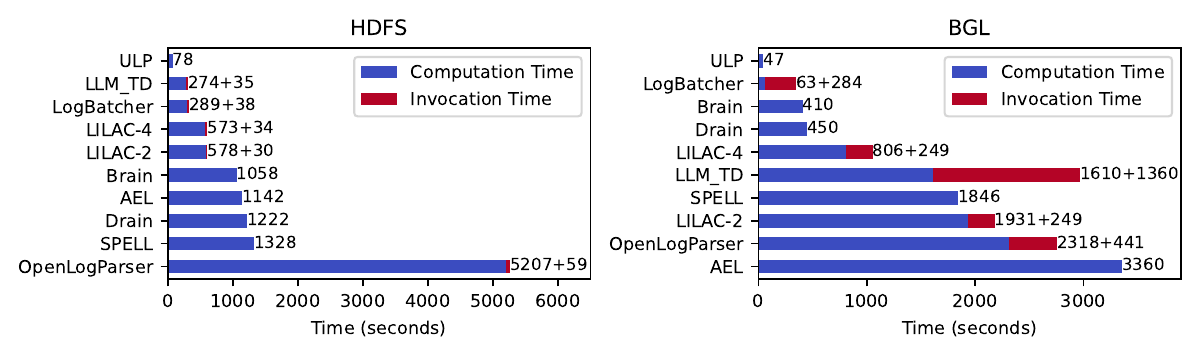}
    \caption{Computation and invocation time of parsers on HDFS and BGL with GPT-3.5.}
    \label{fig:time}
\end{figure}

\begin{figure}[h]
    \centering
    \includegraphics[width=\linewidth]{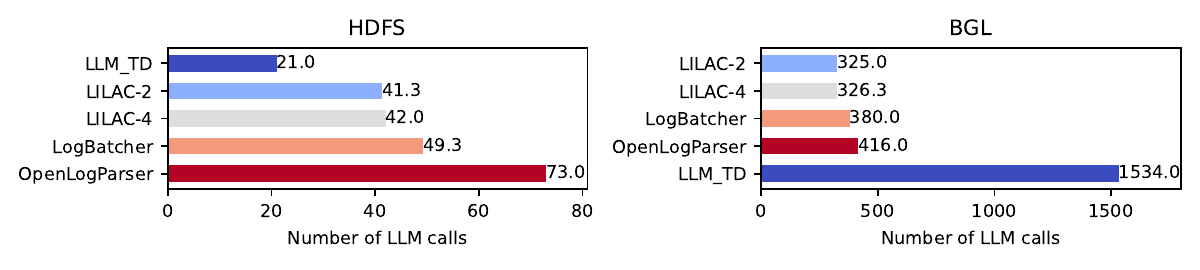}
    \caption{Number of LLM calls of parsers on HDFS and BGL with GPT-3.5.}
    \label{fig:queries}
\end{figure}

Figure \ref{fig:time} visualizes the computation time and LLM invocation time of the parsers, Fig. \ref{fig:queries} shows the number of LLM calls made. The invocation time for conventional parsers is, of course, zero. Outstanding runtime efficiency is achieved by the conventional parser ULP \cite{sedki_effective_2022} on both datasets. While LLM-TD is the second fastest and also calls the LLM the least times it is rather the opposite on the BGL dataset. As mentioned in Sec. \ref{sec:code-quality} and \ref{sec:code-changes} there were some implementation problems with SelfLog \cite{pei_self-evolutionary_2024}. They provided a faulty script for the efficiency evaluation which is why we ran the evaluation with the script they designed for the evaluation of the $2000$ log lines version of LogHub \cite{zhu_loghub_2023}. However, for both HDFS and BGL the processes got killed and are therefore not featured in the plots.

\section{Discussion}\label{sec:discussion}
This section summarizes the findings of the literature review from Sec. \ref{sec:literature-review-results} and of the evaluation from Sec. \ref{sec:evaluation-results} by answering the research questions from Sec. \ref{sec:intro}.

\subsection{Answers to Research Questions}

\subsubsection*{RQ1: What are the main advantages and disadvantages of LLM-based log parsing approaches and non-LLM-based approaches?}
LLM-based log parsing approaches offer key advantages such as adaptability to diverse log formats, the ability to generalize across different system logs, and robustness to unseen log templates as demonstrated by the evaluation on the Audit dataset \cite{landauer_maintainable_2023, landauer_have_2021} in Sec. \ref{sec:audit-evaluation} and other work \cite{fu_investigating_2022}. Approaches like the unsupervised parser LogBatcher \cite{xiao_demonstration-free_2024} and the supervised parser LILAC \cite{jiang_lilac_2024} can handle complex and unstructured logs more effectively than conventional methods, as demonstrated in Sec. \ref{sec:baseline-performance} and Sec. \ref{sec:LLM-parsers-performance} and their corresponding publications \cite{jiang_lilac_2024, xiao_demonstration-free_2024}. They utilize both syntactic and semantic information within log data, outperforming methods that focus solely on syntactic (e.g. Drain \cite{he_drain_2017}, SPELL \cite{du_spell_2016}) or semantic information (e.g. DivLog \cite{xu_divlog_2024}). However, they also come with drawbacks. The usage of LLMs comes with a significant computational burden, implying the requirement of hardware capable of running LLMs locally or using external API services. Methods like caching can significantly reduce the number of LLM calls and therefore also the latency, but especially systems with a high number of templates or frequently changing templates (e.g. due to updates) may still use too many resources for certain use cases. The use of LLMs also introduces randomness into the process due to the randomness within the LLM's output and possibly overthinking for reasoning models \cite{cuadron_danger_2025}. In contrast, approaches that do not integrate language models into their process, such as Brain \cite{yu_brain_2023} and ULP \cite{sedki_effective_2022}, have deterministic output, are computationally efficient, and are often easier to deploy but lack the flexibility of LLMs in handling diverse log templates.

\subsubsection*{RQ2: To what extent do LLM-based log parsing approaches rely on labeled data and manual configuration?}
LLM-based log parsing approaches typically require labeled datasets for fine-tuning or evaluation, though some methods leverage self-evolutionary approaches using previously parsed templates to substitute the need for a priori labels. However, with only $2$ and $4$ shots of labeled logs, LILAC \cite{jiang_lilac_2024} achieves superior performance on the corrected LogHub dataset \cite{khan_guidelines_2022} and Audit \cite{landauer_maintainable_2023, landauer_have_2021}, compared to the conventional parsers and other LLM-based parsers, excluding LogBatcher \cite{xiao_demonstration-free_2024} (Sec. \ref{sec:evaluation-results}). Furthermore, LogBatcher demonstrates impressive performance on both runtime and effectiveness while being unsupervised and only requiring the log format as input parameter. Fine-tuning \cite{karanjai_logbabylon_2024, ma_llmparser_2024, mehrabi_effectiveness_2024, pang_large_2024, zhong_logparser-llm_2024} or even pretraining approaches \cite{ji_adapting_2024} also report outstanding performance, yet their stronger reliance on labeled logs and computational power and the robust and high performance of LogBatcher and LILAC demonstrate that ICL is a cheaper but still performant alternative.

Compared to the baseline parsers AEL \cite{jiang_abstracting_2008}, SPELL \cite{du_spell_2016}, Drain \cite{he_drain_2017}, ULP \cite{sedki_effective_2022} and Brain \cite{yu_brain_2023}, the LLM-based parsers of our evaluation selection require on average less configuration parameters (Sec. \ref{sec:manual-configuration}). This finding is particularly noteworthy in light of the satisfactory performance achieved by LogBatcher \cite{xiao_demonstration-free_2024}. This finding suggests that the deployment of LLMs can effectively substitute a substantial proportion of manual effort, and thus enhance the usability.

\subsubsection*{RQ3: Which techniques can enhance the efficiency or effectiveness of LLM-based log parsing?}
It is evident that ICL and fine-tuning present powerful learning paradigms for LLM-based log parsing. They are not mutually exclusive to each other and can be employed simultaneously to improve parsing accuracy (Sec. \ref{sec:learning}). ICL can be enhanced with RAG or with previously parsed or labeled logs as demonstrations. The ablative studies of reviewed papers \cite{jiang_lilac_2024, xu_divlog_2024, xiao_demonstration-free_2024, xu_help_2024, zhang_eclipse_2024} unanimously conclude that dynamic demonstrations and sophisticated demonstration selection algorithms improve accuracy scores (Sec. \ref{sec:RAG}). Smart caching solutions show efficiency improvements in terms of runtime but also monetary expenses since LLM calls are usually costly (Sec. \ref{sec:caching}). Template revision methods that correct, delete, or merge related templates in postprocessing can further improve cached templates. This can also help adapt the parser to evolving logs due to updates or other behavioral changes of the computer systems without reconfiguration or retraining (Sec. \ref{sec:template-correction}). Evaluations show that it is also possible to improve the effectiveness of parsers with more powerful LLMs, yet the effective application of reasoning models like DeepSeek R1 should be investigated more closely (Sec. \ref{sec:evaluation-results}). It has been shown that variable-aware prompts improve the effectiveness in direct parsing with LLMs. Naturally, more explanations in the prompt mean more tokens per parsed log, which negatively affects the inference time of LLMs. Short and concise instructions and explanations could thus alleviate this issue. Furthermore, creating coarse-grained clusters to capture the commonalities of logs and fine-grained clusters to capture the variabilities of logs to support the template extraction process constitutes another promising technique, that can be applied by conventional and LLM-based parsers alike (Sec. \ref{sec:LLM-Usage}).

\subsubsection*{RQ4: Which experiment design choices hinder the comparability, and hence which guidelines should be adopted in terms of configuration, datasets used, evaluation metrics, and reporting to make results comparable?}
Consistency in preprocessing methods is crucial for fair comparisons. For instance, not all parsers require the log format (see definition in Sec. \ref{sec:background}) as an input parameter, such as LLM-TD \cite{vaarandi_using_2024} and LUNAR \cite{huang_lunar_2024}. LUNAR explicitly includes log headers, such as IP addresses, timestamps, or levels, to grasp the full context of the log. For instance, partitioning logs into groups based on their timestamps is also possible with time intervals \cite{zhou_leveraging_2024}. Previous research \cite{he_evaluation_2016} found that parsing only the extracted log message part, according to the log format, improves effectiveness, but especially LLMs with their generalization capabilities could utilize this extended context \cite{huang_lunar_2024}. In general, the choice of configuration parameters significantly impacts comparability. For example, the researchers of Lemur \cite{zhang_lemur_2025} report near-perfect scores of $0.999$ and $0.996$ for FGA and GA on LogHub \cite{zhu_loghub_2023}, yet they require log format, regular expressions, and four other hyperparameter-tuned parameters for each dataset. This demonstrates the potential performance levels that can technically be achieved. However, the number of parameters required is impractical for real-world applications and complicates meaningful performance comparisons.

The datasets related to LogHub \cite{zhu_loghub_2023, jiang_abstracting_2008, khan_guidelines_2022} are used in most publications, namely $25$ out of $29$. LogHub and its versions are widely used benchmarks, enabling direct comparison with a large body of existing research and providing a standardized way to evaluate parsing performance. We therefore recommend using at least one of the LogHub versions for the sake of comparability. However, relying solely on LogHub datasets may not cover all possible log structures and scenarios. Incorporating custom or other open-source datasets reveals how well parsers generalize to different types of logs, identifies their strengths and weaknesses, and avoids overfitting to a single dataset (or dataset collection) (Sec. \ref{sec:datasets}).

We recommend a set of metrics, namely GA, PA, FTA, and NED, to be standardized across studies to cover the relevant characteristics of templates, since each metrics maps to specific requirements of specific downstream tasks (Sec. \ref{sec:metrics}). This is important to prevent single-sided metric selection that focuses on single characteristics. Using FGA over GA should also be considered if class imbalance is a concern, but if not, there is little difference whether GA or FGA is used as long as the aspect of grouping performance is covered. The results of our evaluation in Sec. \ref{sec:evaluation-results} show high variance between these metrics for different parsers, indicating that this selection covers a variety of aspects.

The aforementioned experiment design choices concern all types of log parsers. Specific to LLM-based parsers, the LLM employed is also an important factor to consider regarding the comparability of evaluation results. As demonstrated, CodeLlama, GPT-3.5, and DeepSeek R1 perform significantly different on the effectiveness scores (Sec. \ref{sec:evaluation-results}). Additionally, variations in API versions, underlying model updates, and differences in temperature settings further contribute to comparability issues in performance across studies.

\subsubsection*{RQ5: To what extent are the presented LLM-based parsing methods accessible and reproducible?}
The accessibility and reproducibility of the presented LLM-based parsing methods are significantly hindered by issues related to code availability, documentation, and execution reliability (Sec. \ref{sec:code-quality}). Many approaches do not provide public code, and even when code is available, it often lacks essential components, such as necessary scripts or instructions, making replication difficult. Several repositories suffer from missing files, incorrect implementations, or incomplete functionality that does not align with the descriptions in their respective papers. Furthermore, discrepancies in the code, such as missing scripts for crucial steps or incorrect default settings, complicate reproducibility. Even when modifications are made to ensure compatibility --- such as fixing typos, updating libraries, or implementing minor adjustments to improve consistency (Sec. \ref{sec:code-changes}) --- these changes highlight the necessity of external intervention to make the methods functional. The need for such corrections aligns with broader findings from Trisovic et al. \cite{trisovic_large-scale_2022} and Olzewski et al. \cite{olszewski_get_2023}, who observed that a majority of replication code repositories contain errors that prevent immediate execution. Olzewski et al. further found that only a small proportion of the running codebases also produce the claimed results of the related papers \cite{olszewski_get_2023}. Additionally, some parsers require substantial configuration efforts, further limiting accessibility (Sec. \ref{sec:manual-configuration}). Whilst minor improvements, such as adding caching mechanisms to DivLog \cite{xu_divlog_2024} or modifying LogPrompt’s \cite{liu_interpretable_2024} prompt size, enhance usability, they also introduce deviations from the original implementations, raising concerns about reproducibility. The discrepancy between our own evaluation results (Sec. \ref{sec:evaluation-results}) and the reported results raises further concerns about the validity of the publications' evaluations but naturally also about our own evaluation. Future benchmarks are expected to address these concerns. Finally, while log parsers based on LLM show great potential, their implementations' current state significantly hinders accessibility and reliable reproduction without substantial external effort.

\subsection{Threats to Validity}
We identified two major threats to the validity of the results of our evaluation concerning the adoption of the code and randomness.

\subsubsection{Mistakes in Code Adoption}
In Sec. \ref{sec:code-quality} we describe issues with the code of some of the parsers and in Sec. \ref{sec:code-changes} we describe the changes we made due to the mentioned issues and to ensure transparency. We found that a significant number of parsers do not attain the performance levels claimed in their respective papers. This raises the question whether the observed discrepancies can be attributed to our alterations in their code or the inaccuracy of their evaluation. Alternatively, it is possible that other factors, such as the utilization of different language models, are responsible for the observed variations. However, given that the alterations were predominantly minor, we anticipate that the correctness was not significantly negatively influenced. The code has been made available on GitHub for reproduction. Should any flaws in the adoption of the parsers be identified, we request to report an issue.

\subsubsection{Randomness}
The validity of results is threatened by the randomness in LLM output and the randomness introduced by the random sampling for the supervised parsers. To mitigate the randomness in LLMs, we set the LLMs' initial temperature to $0$. The temperature determines the creativity of an LLM. However, even a temperature of $0$ does not guarantee a deterministic output \cite{astekin_exploratory_2024}. Therefore, we repeated the evaluation three times and computed the average of the measured values.

\subsubsection{Data Leakage}
Following previous work \cite{xiao_demonstration-free_2024, jiang_lilac_2024} we identify data leakage through memorized logs from the pretraining phase of the LLMs as a threat to the validity of our evaluation results. However, \cite{jiang_lilac_2024, xu_divlog_2024, xiao_demonstration-free_2024} report significant performance improvements with ICL compared to direct zero-shot parsing, suggesting a low likelihood of GPT-3.5-Turbo memorizing single log templates. In a small experiment we asked CodeLlama and DeepSeek R1 if they can output exact logs of the LogHub repository. In both cases they were not able to print exact logs, that matched any of those from the repository. The same held true when asked for templates. Diverse studies \cite{schwarzschild_rethinking_2024, lu_scaling_2024} suggest that direct fact memorization is unprobable and that the models are only likely to memorize often encountered text from diverse sources from their training \cite{schwarzschild_rethinking_2024}.

\section{Conclusion}\label{sec:conclusion}

In this work, we conducted a systematic review of LLM-based log parsing approaches, analyzed their strengths and weaknesses in comparison to non-LLM-based methods, and performed an extensive benchmark evaluation to assess their effectiveness, efficiency, and usability. Our findings demonstrate that LLM-based log parsers provide notable advantages, particularly in their adaptability to diverse log formats and their capability to generalize across unseen templates. However, they also come with challenges such as high computational costs, susceptibility to hallucinations, and issues with interpretability.

A key observation is that LLM-based log parsers often reduce the need for manual configuration and labeling, making them more accessible to users with limited domain expertise. Techniques such as in-context learning (ICL), fine-tuning, retrieval-augmented generation (RAG) and template revision can significantly enhance efficiency and accuracy of these methods. Sophisticated dynamic demonstration selection and caching strategies have been shown to significantly impact performance, reducing both LLM inference time and cost. However, despite these improvements, only two out of seven LLM-based parsers, namely LILAC \cite{jiang_lilac_2024} and LogBatcher \cite{xiao_demonstration-free_2024}, clearly outperform the non-LLM-based parsers that constitute our baseline. Furthermore, we identified a lack of reproducibility and comparability, as many implementations lack publicly available code or datasets, comprehensive documentation, or consistency with reported results. Our study highlights the necessity of standardized benchmarking practices for LLM-based log parsing. Differences in experimental setups, dataset preprocessing, and metric selection complicate direct comparisons between methods and impede a meaningful selection for user-specific downstream tasks.

Overall, LLM-based log parsing represents a promising direction for automated log analysis. Techniques like caching, RAG, template revision, especially when used in combination, have clearly proven their value in this field, but further research is required to fully realize the potential of LLM-based log parsing. We hope that our systematic review, benchmark study, and insights contribute to the development of more effective, transparent, and user-friendly log parsing solutions. To facilitate future research and reproducibility, we make our evaluation results and source code publicly available on our GitHub repository \cite{beck_github_2025}.

\begin{acks}
Funded by the European Union under the European Defence Fund (GA no. 101121403 - NEWSROOM) and under the Horizon Europe Research and Innovation programme (GA no. 101168144 - MIRANDA). Views and opinions expressed are however those of the author(s) only and do not necessarily reflect those of the European Union or the European Commission. Neither the European Union nor the granting authority can be held responsible for them. This work is co-funded by the Austrian FFG Kiras project ASOC (GA no. FO999905301).
\end{acks}

\bibliographystyle{ACM-Reference-Format}
\bibliography{references}

\end{document}